\documentclass[letterpaper]{article} 
\usepackage{aaai24} 
\usepackage{times}  
\usepackage{helvet}  
\usepackage{courier}  
\usepackage[hyphens]{url}  
\usepackage{graphicx} 
\usepackage{natbib}  
\usepackage{caption} 
\usepackage{algorithm}
\usepackage{algorithmic}

\usepackage{newfloat}
\usepackage{listings}
\DeclareCaptionStyle{ruled}{labelfont=normalfont,labelsep=colon,strut=off} 
\floatstyle{ruled}
\newfloat{listing}{tb}{lst}{}
\floatname{listing}{Listing}
\usepackage{amsmath}
\usepackage{amssymb}
\usepackage{mathtools}
\usepackage{amsthm}

\usepackage{microtype}
\usepackage{graphicx}
\usepackage[section]{placeins} 

\nocopyright 

\usepackage{subcaption}
\usepackage{booktabs} 
\usepackage[capitalize,noabbrev]{cleveref}

\title{Curriculum Learning with Adam: The Devil Is in the Wrong Details}
\author {
    Lucas Weber,\textsuperscript{\rm 1}
    Jaap Jumelet,\textsuperscript{\rm 2}
    Paul Michel,\textsuperscript{\rm 3}
    Elia Bruni,\textsuperscript{\rm 4}
    Dieuwke Hupkes\textsuperscript{\rm 5}
}
\affiliations {
    \textsuperscript{\rm 1} DTCL, University Pompeu Fabra, Barcelona, Spain\\
    \textsuperscript{\rm 2} ILLC, University of Amsterdam, Amsterdam, Netherlands\\
    \textsuperscript{\rm 3} CSD, École normale supérieure PSL, Paris, France\\
    \textsuperscript{\rm 4} IKW, Osnabrück University, Osnabrück, Germany\\
    \textsuperscript{\rm 5} FAIR\\
    lucas.weber@upf.edu, j.w.d.jumelet@uva.nl, pmichel31415@gmail.com, elia.bruni@gmail.com, dieuwkehupkes@meta.com
}

\begin{document}

\maketitle

\begin{abstract}
Curriculum learning (CL) posits that machine learning models -- similar to humans -- may learn more efficiently from data that match their current learning progress.
However, CL methods are still poorly understood and, in particular for natural language processing (NLP), have achieved only limited success.
In this paper, we explore why.
Starting from an attempt to replicate and extend a number of recent curriculum methods, we find that their results are surprisingly brittle when applied to NLP.
A deep-dive into the (in)effectiveness of the curricula in some scenarios shows us why: when curricula are employed in combination with the popular Adam optimisation algorithm, they oftentimes learn to adapt to suboptimally chosen optimisation parameters for this algorithm.
We present a number of different case studies with different common hand-crafted and automated CL approaches to illustrate this phenomenon, and we find that none of them outperforms optimisation with only Adam with well-chosen hyperparameters.
As such, our results contribute to understanding why CL methods work, but at the same time urge caution when claiming positive results.

\end{abstract}

\section{Introduction}
\label{sec:intro}
State-of-the-art machine learning is becoming increasingly computationally expensive. 
In the interest of saving resources and making the latest innovations more accessible to the broader public, there is a strong research interest in more efficient learning methods.
A popular approach to data-efficient learning is curriculum learning (CL; see \citealt{elman1993learning, bengio2009curriculum};  see also \citealt{soviany2022curriculum, wang2021survey} for reviews), which suggests that models -- similar to humans -- learn optimally from data that match their current learning progress.

While curriculum learning has been successful in certain research areas \citep[most notably in reinforcement learning; ][]{narvekar2020curriculum}, it had mixed success in the field of natural language processing (NLP). 
In a very common setting of state-of-the-art NLP -- consisting of language model pretraining and subsequent fine-tuning -- curriculum learning has seen no success in the pretraining stage \citep[e.g.][]{surkov2021data, campos2021curriculum} and only produced marginal improvements in the fine-tuning stage \citep[e.g.][]{xu2020curriculum}.
Due to these mixed results, there are no simple out-of-the-box solutions or clear guidelines for the use of curricula in NLP and they, therefore, find little use.
In this paper, we conduct an empirical analysis of curriculum learning and come to the conclusion that mixed results in NLP might be related to the widespread use of the Adam optimiser \citep{kingma2014adam} in the field.

We start our analysis by conducting a case study on an existing fully-automated curriculum learning approach \citep{raghu2020teaching} from computer vision (Section~\ref{sec:case_study_commentaries}). 
We reproduce the results of the original paper, show how it can also produce learning advantages on language data and that its policy is in line with other successful approaches from the existing literature (Section~\ref{subsec:comm_exp_results}).
Despite these apparently sound results, we also find the approach to be brittle and inconsistent.
Upon deeper investigation, we find that, rather than providing a sound data-based curriculum strategy, the learned curricula are fully data-agnostic and stem from interactions of the curriculum shape with the Adam optimiser, rather than a sound curriculum strategy (Section~\ref{sec:cl_adam_interaction}).
As a result of the interaction, the parameter updates of the model are scaled in size, similar to a change in learning rate.
Similar or larger learning advantages can be achieved by properly tuning hyperparameters.

In a second set of experiments, we go on to demonstrate how the \emph{curriculum-Adam}-interaction is not limited to the commentaries framework.
We will lay out in Section~\ref{sec:background} how all curriculum learning approaches share a similar structure. %
We, therefore, continue to test common, simple hand-crafted curricula and, here, observe interactions as well.
Importantly, plain Adam with properly tuned hyperparameters outperforms curricula in all of our tested settings.

\begin{figure*}
  \centering

    \includegraphics[width=0.65\linewidth]{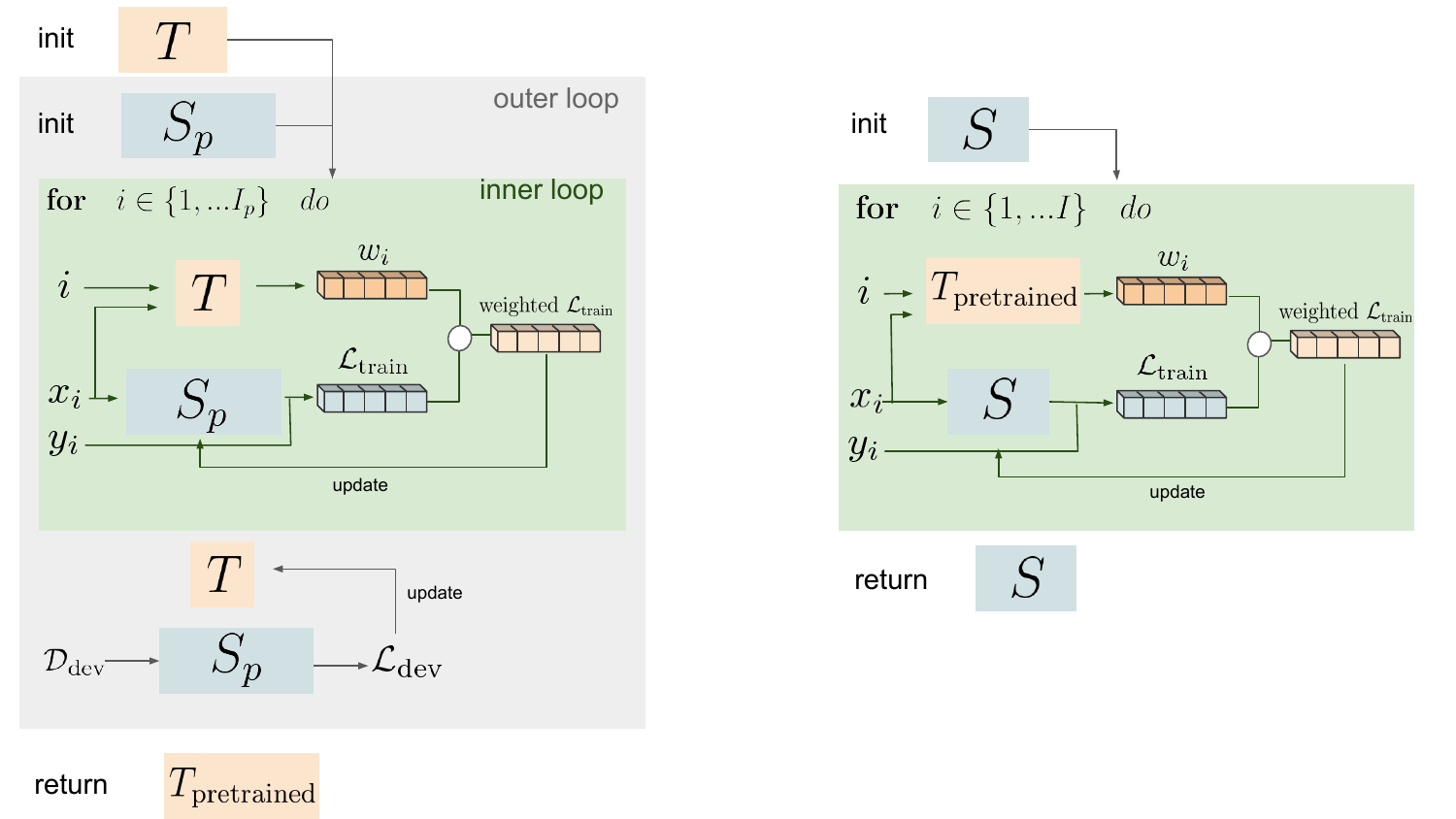}

  \caption{A visualisation of the commentaries-framework. The left side illustrates the teacher optimisation: The teacher model ($T$) is trained in the outer loop to optimise the learning process of the practice target models ($S\textsubscript{p}$) in the inner loop. 
  The number of iterations in the inner loop is limited by the amount of memory available. 
  The right side shows how the pretrained teacher is used to optimise a new student model to convergence.}
  \label{fig:commentary_algorithm}
\end{figure*}

We can summarise our contributions as follows:
\begin{enumerate}
 \item We transfer commentaries -- an automated curriculum learning approach \citep{raghu2020teaching} -- from vision- to language data and provide an empirical analysis of its behaviour.
\item We showcase how commentaries work not due to a beneficial ordering of the data, but rather by a data-agnostic interaction with the optimiser. This can fully explain the learning advantages attributed to the curriculum.
\item We expand the notion to other types of curricula that are commonly used with Adam and empirically demonstrate how these curricula are affected as well.
\end{enumerate}


\section{Background}
\label{sec:background}

Inspired by human learning, curriculum learning (CL) exposes machine-learning models to a limited, `simple' portion of the data distribution at first and only gradually introduces `complex' examples into the training process until the whole training data is used \citep[][]{elman1993learning, rohde1999language, krueger2009flexible, bengio2009curriculum}.  
To this end, every CL approach has to formalise which training examples are `simple' and which are `complex' (i.e. determine a \textit{difficulty measure}) and decide on the rate at which to add `more complex' examples into training (i.e. define a \textit{scheduling function}).
Difficulty measures and schedule functions can be determined in different ways.
We here shortly summarise a broad grouping of approaches: \textbf{hand-crafted curricula} and \textbf{automated curricula}.

\paragraph{Hand-crafted curricula}
The simplest type of curriculum fixes the difficulty measure and schedule function prior to training, without adapting them dynamically according to the learner state.
The choice of the difficulty measure is usually based on the practitioner's intuitions and experiences. 
Common \emph{difficulty measures} in NLP include the sequence lengths of an input (or the closely related depth of the parse tree) \citep{tay2019simple, alonso2017parsing, platanios2019competence}, the number of coordinating conjunctions \citep{kocmi2017curriculum} or the diversity of the used vocabulary \citep{platanios2019competence}.
\emph{Schedule functions} typically expand the data distribution towards more difficult examples monotonically, either as a step-function \citep{bengio2009curriculum, spitkovsky2010baby, kocmi2017curriculum} or continuously \citep{hacohen2019power, platanios2019competence, penha2020curriculum, liu2018curriculum}.
Examples of step functions can be seen in Figure~\ref{subfig:schedule_functions}.
Hand-crafted curricula have the advantage of being cheap and easy to implement.
On the other hand, the choice of the correct setup requires experience or expert domain knowledge, idiosyncracies of data and tasks make them potentially difficult to generalise and the method is `coarse', such that it is limited to the predefined structure and cannot dynamically adapt to the current state of the learner.

\paragraph{Automated curricula}
There are different approaches to addressing the shortcomings of hand-crafted curricula.
We coarsely bin them into (1) non-parametric and (2) parametric solutions.
The (1) non-parametric curricula can dynamically adapt the schedule function and/or difficulty measures to the current state of the learner without learning any additional parameters. 
The most common approach to non-parametric curriculum learning is self-paced learning \citep[SPL;][]{kumar2010self}. 
In SPL, data points are only included in training when they produce losses that fall under a dynamic threshold. 
 On the other hand, (2) parametric approaches utilise meta-learning to learn additional parameters (often times referred to as `teacher'-models) that predict a data point's utility towards a target (or `student')-model's  learning objective (for examples see MentorNet by \citealt{jiang2018mentornet}, ScreenerNet by \citealt{kim2018screenernet}, and learning-to-teach by \citealt{fan2018learning}).
The predicted utility is then used to optimise the learning process.
As they require no manual work, end-to-end approaches are convenient. However, they come oftentimes with the high computational cost of optimising `teacher' models, making them too expensive to optimise with large target models.

\begin{figure*}
  \centering
  \begin{subfigure}{0.30\linewidth}
    \includegraphics[width=\linewidth]{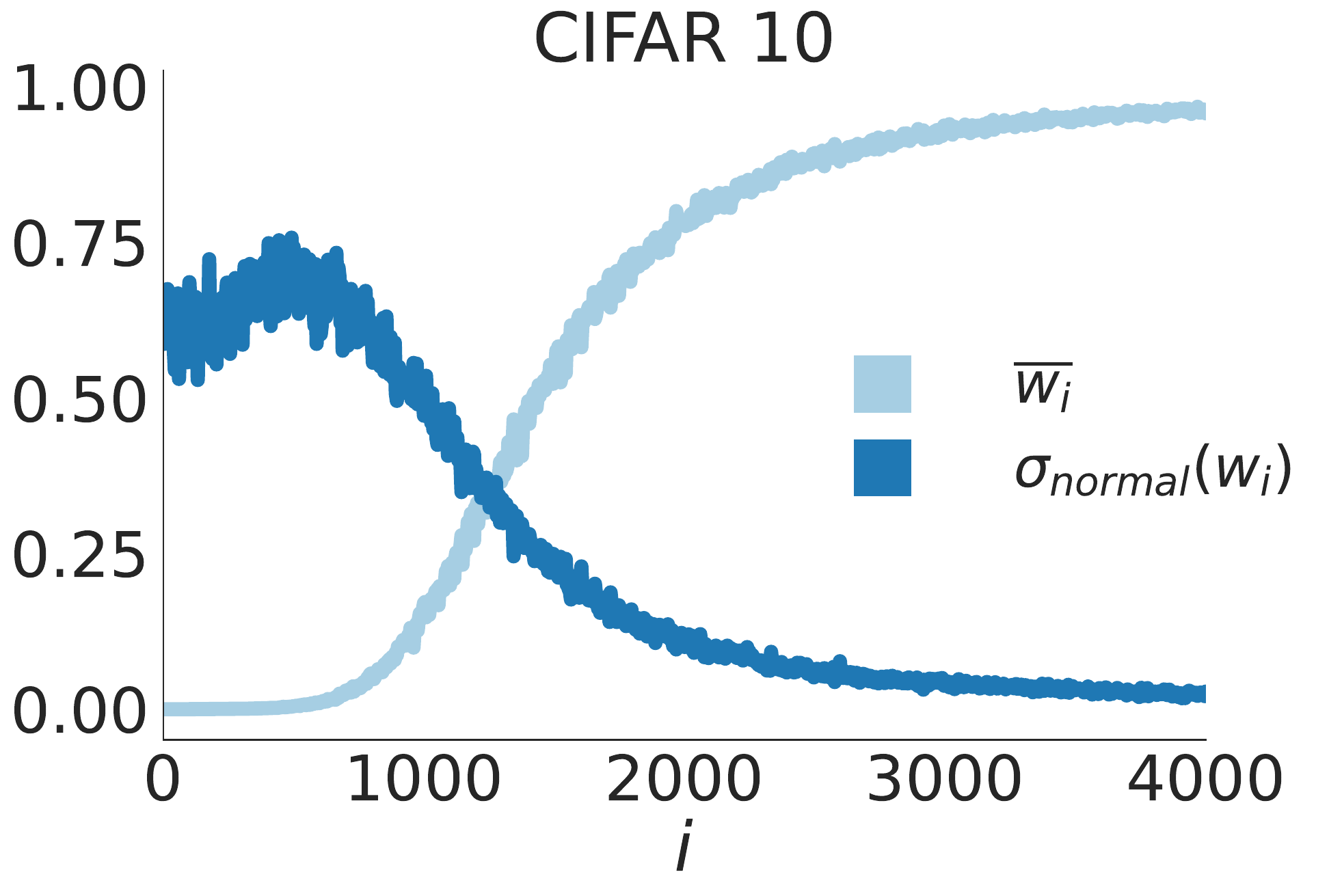}
    \caption{}
    \label{subfig:weight_stats_commentaries}
  \end{subfigure}
  \begin{subfigure}{0.25\linewidth}
    \includegraphics[width=\linewidth]{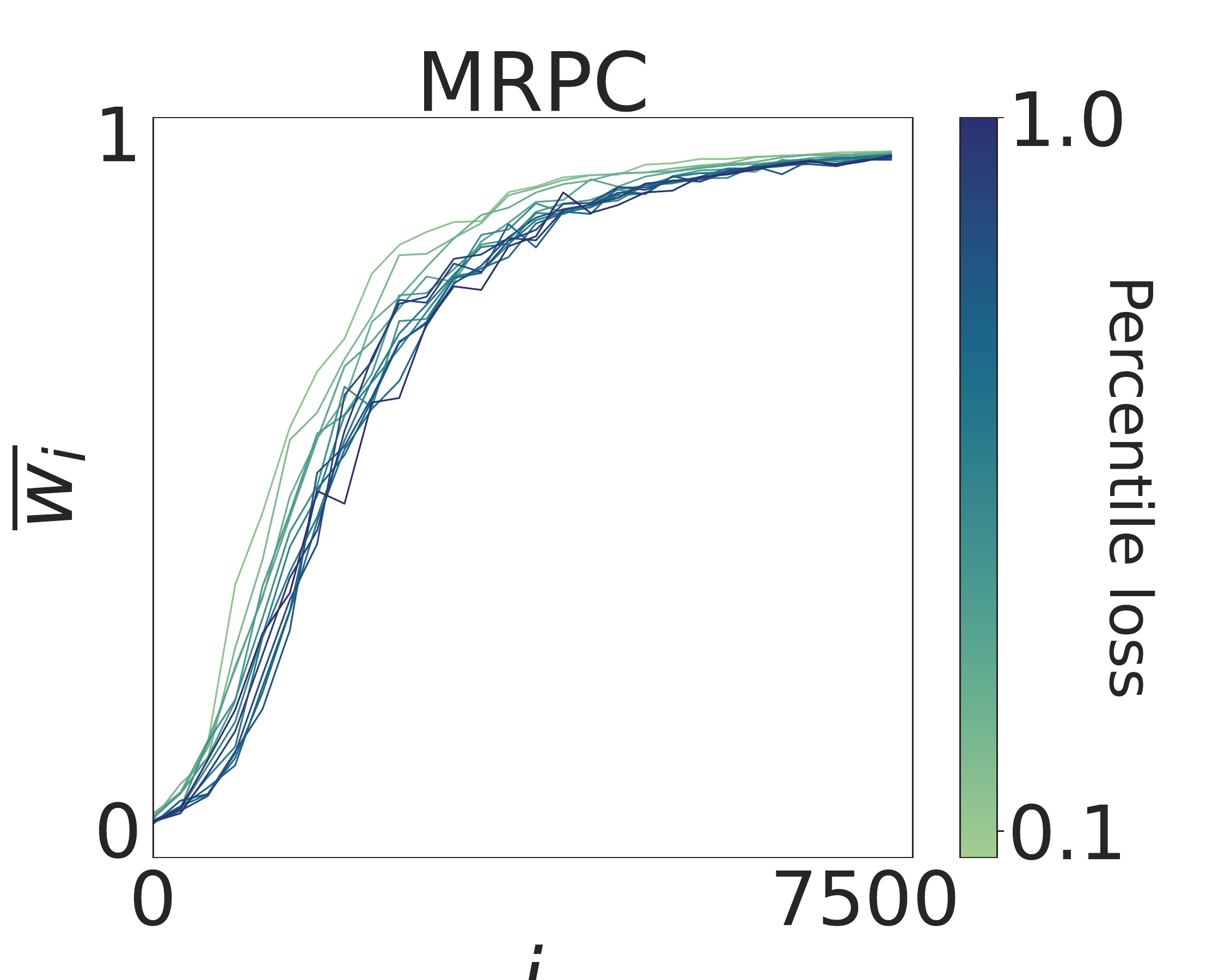}
    \caption{}
    \label{subfig:comms_vs_difficulty_measure_loss}
  \end{subfigure}
  \begin{subfigure}{0.25\linewidth}
    \includegraphics[width=\linewidth]{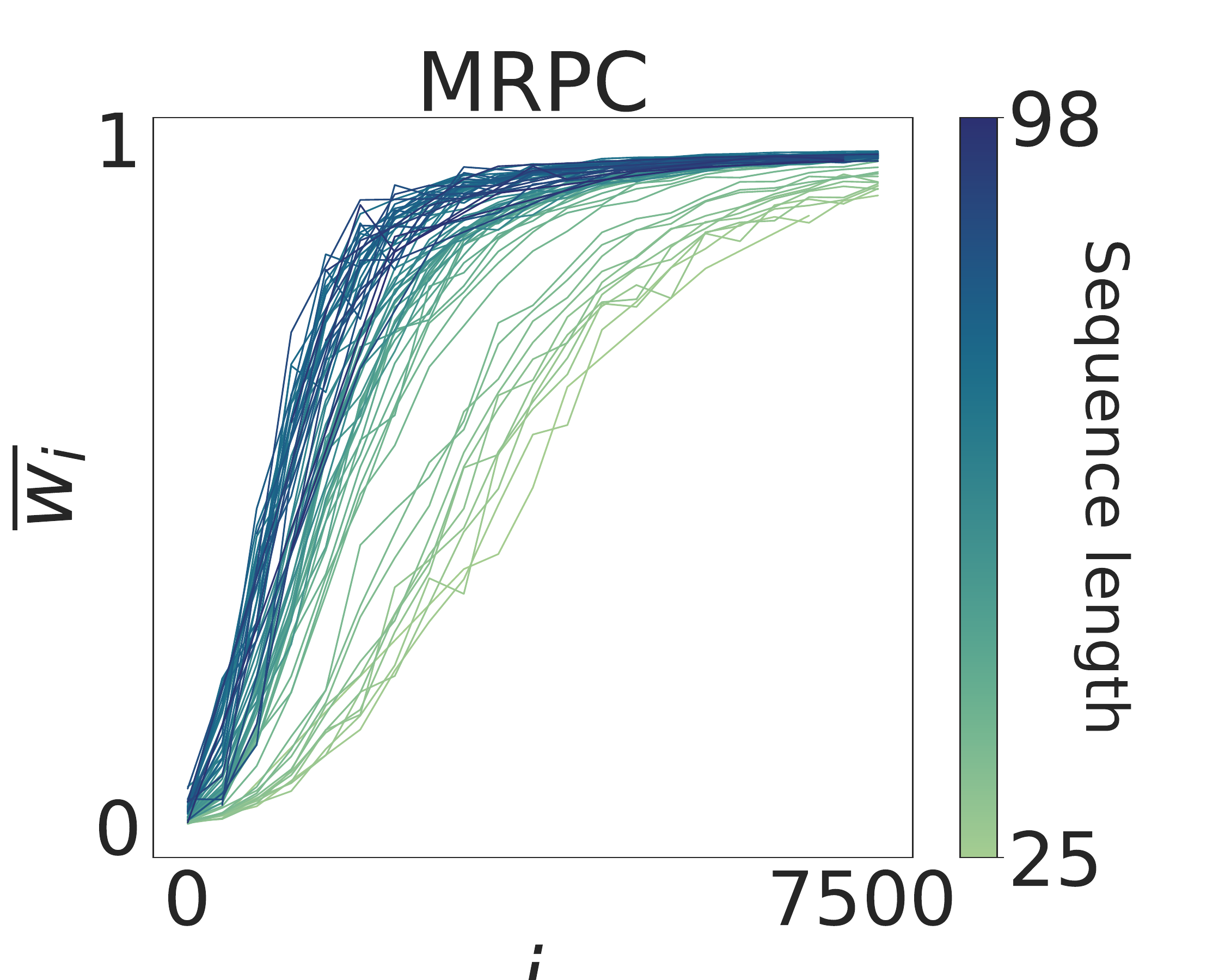}
    \caption{}
    \label{subfig:comms_vs_difficulty_measure_seq_len}
  \end{subfigure}
  \caption{All of our pretrained commentary teachers (vision and language) show the same pattern when predicting weights (a): predicting small value, high-variance weights early in student training to then predict higher and more uniform weights as student training progresses. 
  When trained on an NLU task like MRPC\footnote{We choose MRPC as we consider it representative of most GLUE tasks. We find equivalent results for other GLUE-tasks.} \citep{dolan2005automatically}, the teacher shows a slight preference for training examples with lower loss by assigning higher weights to these examples earlier in training (b). The preference is even clearer for its weighting policy in regard to sequence lengths (c): longer sequences are weighted up the beginning of training and longer sequences later are only included later. Loss and sequence length are common difficulty measures in CL.}
  \label{fig:comms_vs_difficulty_measure}
\end{figure*}

\paragraph{Theoretical underpinnings}
Theoretical explanations of the efficiency of curriculum learning remain relatively sparse.
The two most referred-to explanations can be found in \citet{bengio2009curriculum} which state that CL helps 1) with denoising the dataset and 2) by smoothening of the non-convex optimisation landscape \citep[as a form of continuation method; compare][]{allgower1980numerical}.

Despite all of their different forms and technical implementations, all curriculum learning approaches have in common that they cause a systematic shift in the learning signal the model is receiving.
We refer to this universal shift of curricula as the \emph{curriculum structure} throughout this paper.
The curriculum structure is central to generalising our findings in later sections.


\section{A case study with Commentaries}
\label{sec:case_study_commentaries}
We here conduct a case study on \textit{commentaries}, an existing parametric approach to curriculum learning.
We start by summarising how the commentaries curriculum \citep{raghu2020teaching} is learned and applied.

\paragraph{Mechanism}
To learn a curriculum, commentaries are formalised as a teacher model
$T(x_i,i;\phi) \rightarrow w_i$ with parameters $\phi$ that takes a batch of data $x_i$ and an indicator of the target model's current learning state $i$ to produce a weight $w_i \in$ [0,1] for every data point in the batch.
The indicator $i$ is set to be the number of previous iterations for which the target model has been trained and we denote $I$ to be the total amount of updates for which we will train a model.
Further, we denote the target model as $S$ and its parameters as $\theta$.
At every iteration $i$, the weight-vector $w_i$ is applied to the target model's loss $\mathcal{L}$\textsubscript{train}. 

The commentaries pipeline is divided into two phases: a teacher-pretraining phase and an evaluation phase.
We depict both phases in Figure~\ref{fig:commentary_algorithm}.
During teacher-pretraining, the teacher is explicitly trained to minimise the loss of $S$ on some held-out data $\mathcal{D}$\textsubscript{dev} by reweighing the training loss of $S$.
To do so,  several `practice' target models $S$\textsubscript{p} are trained on $\mathcal{D}$\textsubscript{train} for a limited amount of steps $I$\textsubscript{p} while their loss $\mathcal{L}$\textsubscript{train} is weighted by the teacher-predicted $w$. 
For all training steps, the computational graph of $S$\textsubscript{p} is maintained.
Subsequently, $S$\textsubscript{p} is evaluated on the held-out set $\mathcal{D}$\textsubscript{dev}. 
Clearly, the resulting loss $\mathcal{L}$\textsubscript{dev} depends  $S$\textsubscript{p}'s optimised parameters $\hat{\theta}$.
At the same time, $\hat{\theta}$ depend on the teacher parameters $\phi$ through the reweighing of $\mathcal{L}$\textsubscript{train} during training, such that:

\begin{equation} \label{eq:comm_optimisation}
	\frac{\partial\mathcal{L}\text{\textsubscript{dev}}}{\partial\phi} = \frac{\partial\mathcal{L}\text{\textsubscript{dev}}}{\partial\hat{\theta}} \times \frac{\partial\hat{\theta}}{\partial\phi}
\end{equation}
	
This makes it possible to backpropagate $\mathcal{L}$\textsubscript{dev} `through training' to update the teacher parameters $\phi$.
The number of $S$\textsubscript{p}'s optimisation steps $I$\textsubscript{p} in the teacher pretraining phase is limited by the amount of memory that can be allocated to store the computational graph. 

In the evaluation phase -- after the teacher parameters $\phi$ have been pretrained -- a new target model $S$ is trained to evaluate the teacher policy. 
Since there is no need to save the computational graph of the training at this stage, there is also no limit to the number of training steps $I$, such that we can now train $S$ to convergence.
For additional details, we refer to \citet{raghu2020teaching}.

\begin{figure*}
  \centering
  \begin{subfigure}{0.22\linewidth}
    \includegraphics[width=\linewidth]{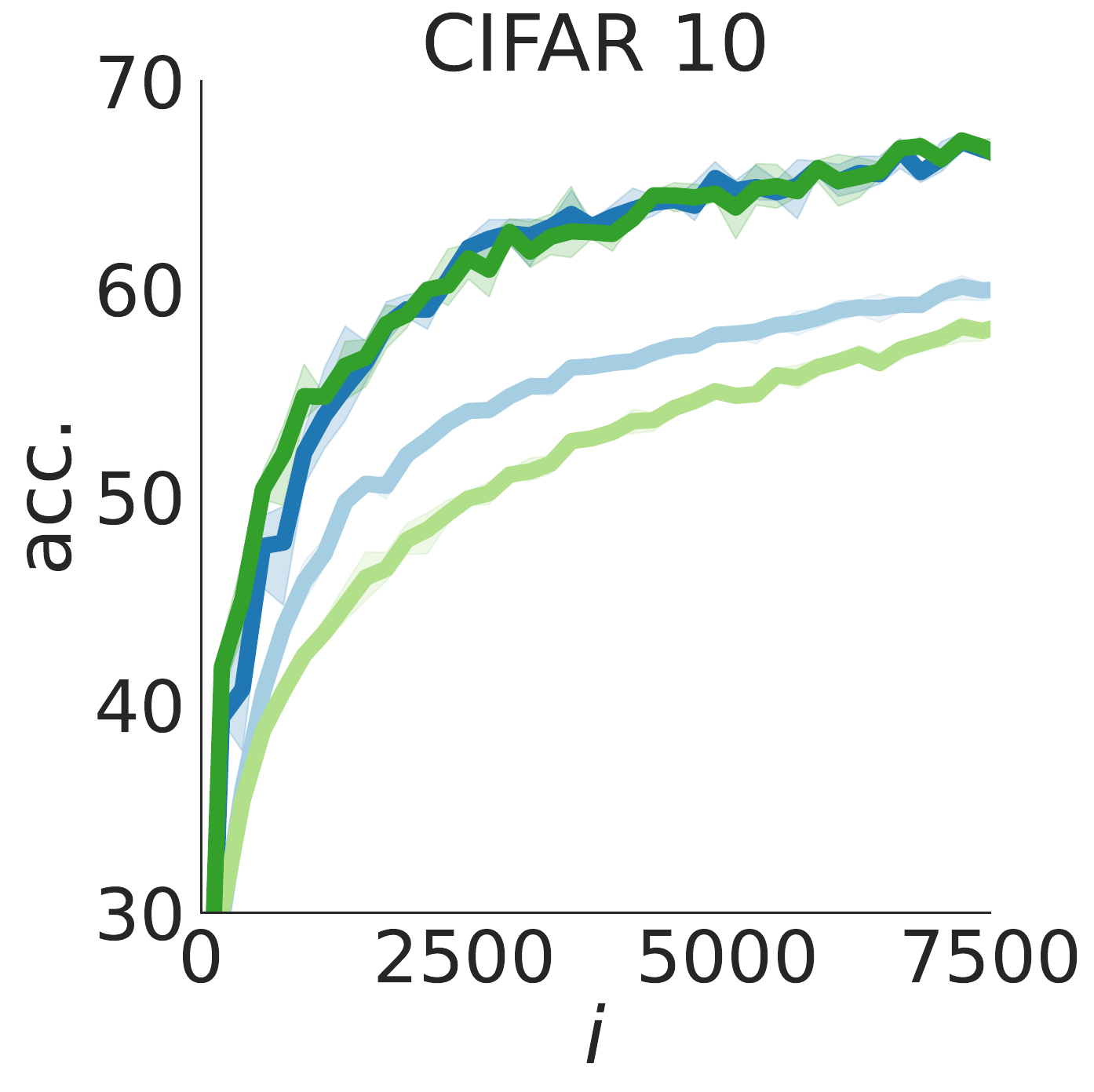}
    \caption{2-layer CNN}
    \label{subfig:replication_cifar10}
  \end{subfigure}
  \begin{subfigure}{0.38\linewidth}
    \includegraphics[width=\linewidth]{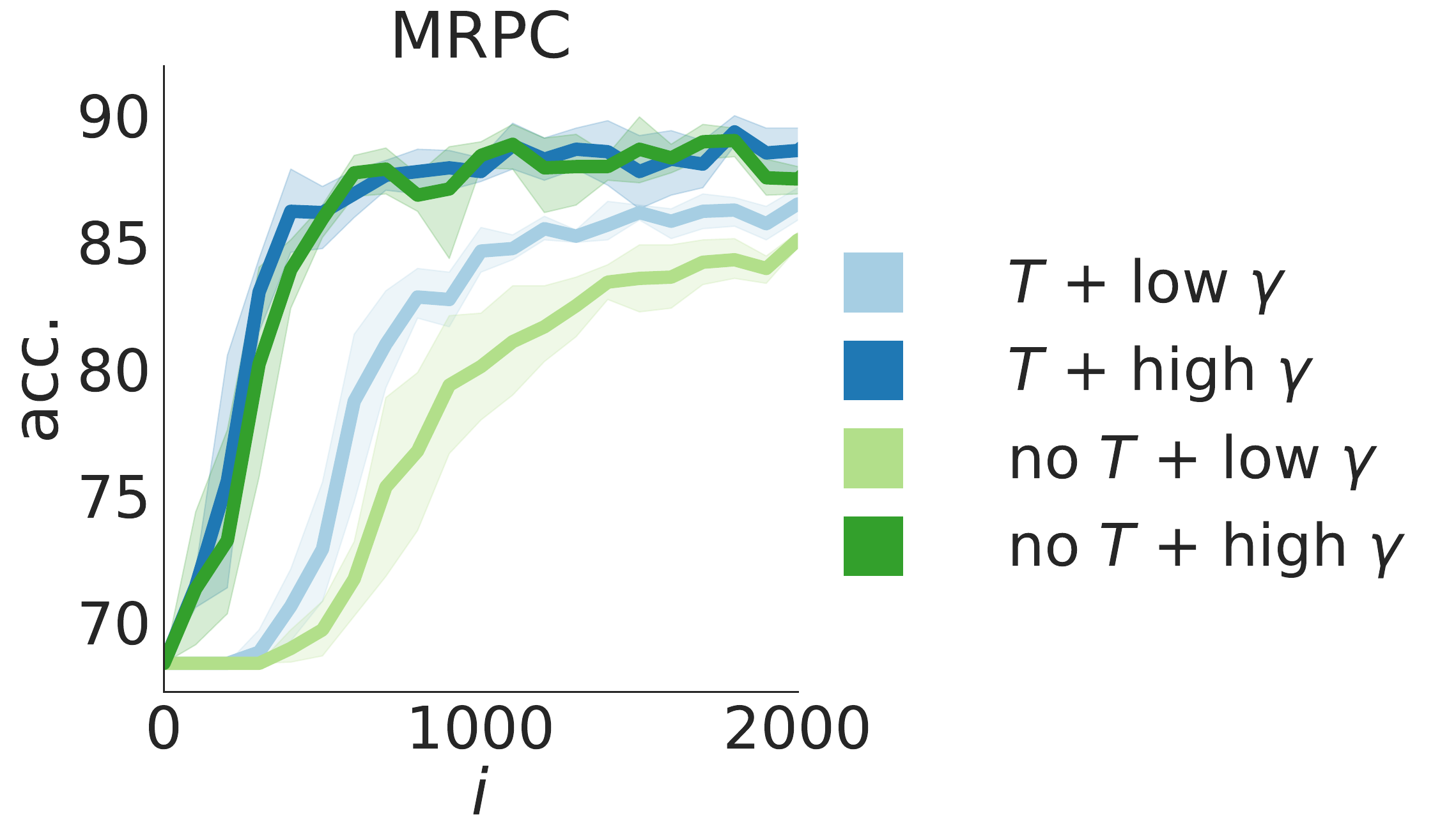}
    \caption{RoBERTa\textsubscript{BASE}}
    \label{subfig:performance_comms_mrpc}
  \end{subfigure}
   \begin{subfigure}{0.27\linewidth}
    \includegraphics[width=\linewidth]{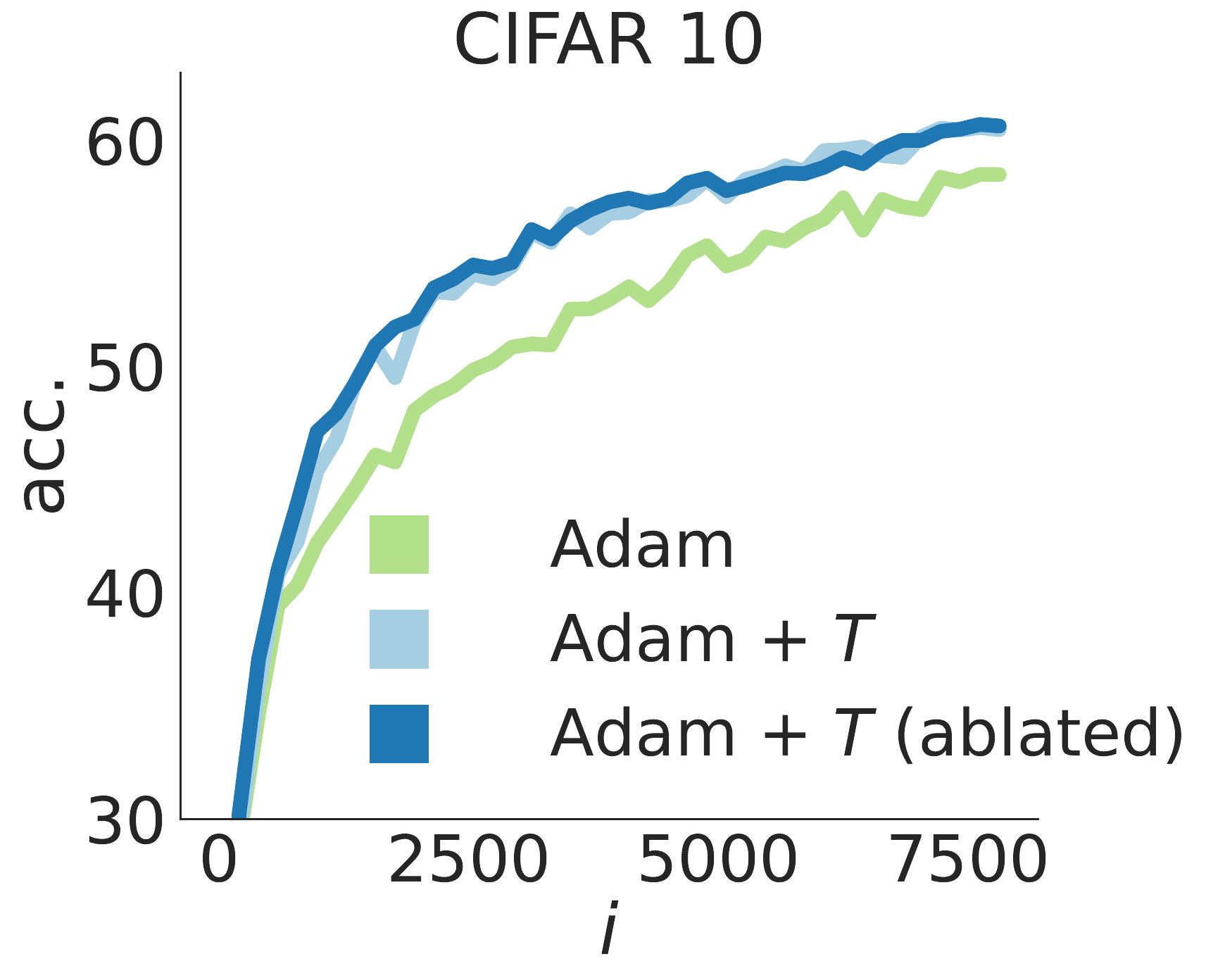}
    \caption{2-layer CNN}
    \label{subfig:org_vs_ablation}
  \end{subfigure}
  \caption{The left side shows a 2-layer CNN (a) trained on CIFAR10 and RoBERTa\textsubscript{BASE} (b) trained on GLUE-MRPC respectively with and without a commentaries teacher ($T$). We see how the teacher achieves learning speed improvements for either model when trained with low learning rates ($\gamma$). However, there is no improvement when hyperparameters are chosen optimally. Figure (c) repeats the low $\gamma$ data from Figure (a) but adds the ablated teacher from Section~\ref{subsec:data_independence} in dark blue for comparison.} %
  \label{fig:replication_experiments}
\end{figure*}

\subsection{Experimental setup}
\label{subsec:comm_exp_setup}
We first replicate \citet{raghu2020teaching}'s results on vision data by using their original code\footnote{\url{https://github.com/googleinterns/commentaries}}.
We train CNN-based teachers with 2-layer CNN-based $S$\textsubscript{p} on the CIFAR10 and CIFAR100 datasets respectively following the previously described procedure while sticking to the reported hyperparameter settings.
After teacher training, we evaluate the teacher on different target models \citep[2-layer CNN, ResNet18, ResNet34; ][]{he2016deep}.

In parallel, we transfer the commentaries framework to language data, specifically to the popular NLU tasks from the GLUE benchmark \citep{wang2018glue}. 
To do so, we replace the CNN-based teacher and target model with small transformer encoder models from the fairseq library \citep[][]{vaswani2017attention, ott2019fairseq}.
To address the computational limitations of the teacher pretraining phase (mentioned in the \emph{mechanism}-paragraph), we use frozen RoBERTa\textsubscript{BASE}-embeddings \citep{liu2019roberta} instead of high-dimensional mappings from the vocabulary as the input to our teacher and target models.
To further reduce the memory requirement of our setup, we average-pool the embeddings with kernel size and stride of 3.
We then optimise teachers with this setup on the GLUE tasks.
We evaluate the teacher by finetuning the full RoBERTa\textsubscript{BASE}-model on the different GLUE tasks with their respective teacher. 

\subsection{Experimental results}
\label{subsec:comm_exp_results}
We first analyse the policy of the teacher models and then continue to evaluate their performance.

\paragraph{Commentaries learn reasonable curricula}
\label{subsubsec:comms_reasonable_curricula}
We first consider the schedule function of the teacher policy.
For CIFAR10, we illustrate in Figure~\ref{subfig:weight_stats_commentaries} how the average weight in every batch $\overline{w_i}$ rises, while the (normalised) standard deviation ($\sigma_{\text{normal}}$) of $w_i$ declines\footnote{Importantly, small weights do not lead to small updates, as Adam normalises the size of the gradient.}.
This means that the teacher model learns a high-variance (i.e.\ selective) weighting in early training which  includes more and more data points as the training of $S$ progresses.
We find a similar policy for the teachers that we trained with language data.
Considering difficulty measures, we find that the teacher's policy is akin to what is known from the literature, making use of sequence lengths \citep{tay2019simple, alonso2017parsing, platanios2019competence} and losses \citep{kumar2010self}.
The teacher schedules long sequences at first and only gradually weighs up short sequences later in training (see Figure~\ref{subfig:comms_vs_difficulty_measure_seq_len}).
Similarly, examples with low losses are introduced first and higher losses are only weighted up afterwards (Figure~\ref{subfig:comms_vs_difficulty_measure_loss}).
The scheduling, as well as the difficulty measures, are in line with what we would expect from the literature (compare Section~\ref{sec:background}).

\paragraph{Commentaries' performance is brittle}
\label{subsubsec:comms_performance_brittle}
We replicate the learning speed improvements that are reported in the original paper (see Figure~\ref{subfig:replication_cifar10}).
In our GLUE setup, we find similar results (Figure~\ref{subfig:performance_comms_mrpc}; for results on other GLUE tasks see Appendix~\ref{app:extension_NLP}). 
However, we also find that these improvements are limited to a certain set of suboptimal hyperparameters. 
As soon as we properly tune the hyperparameters, we learn faster by using the plain Adam optimiser without a teacher (for replication results with all datasets and models see Appendix~\ref{app:replication}). 
In all properly tuned settings, Adam without curriculum performs equally or better. 

In summary, the commentary-teacher's policy very well resembles other successful setups from the CL literature. 
Despite this, we also find that the curricula's benefits during the evaluation phase are surprisingly brittle: Changes in hyperparameters that should not strongly influence the effectiveness of the curriculum -- such as changes in learning rate or batch size -- erase any curriculum advantage.
A proper hyperparameter search makes commentaries ineffective.
Why is this the case, and why are the commentaries working in certain settings, to begin with?
To address these questions, we continue with a more in-depth analysis.

\paragraph{Commentaries are data independent}
\label{subsec:data_independence}
CL assumes that it matters at which point we train on which data point.
We conduct an ablation experiment to see whether this is really what is driving the commentaries' improvements.
We replace the original weighting $w_i$ -- which applies an individual weight for each data point in a batch -- by the batch average $\overline{w_i}$. 
This ablation erases not only the data dependence of the weights but also the distribution of the weights within a batch.
Surprisingly, this ablation does not degrade the curriculum's performance (see Figure~\ref{subfig:org_vs_ablation}) at all.
The exact mapping of data points and weights is thus, apparently, irrelevant.
As a consequence, the learning benefits must originate from the mere shape of the curriculum (or \emph{curriculum structure}), by shifting from small to large weights with increasing $i$.
We corroborate this finding by conducting an additional small experiment with toy curricula that employ different simple weight shifts as their weighting policy: 

\begin{alignat*}{2}
& T_\uparrow\textsubscript{linear}(i) = \tfrac{i}{\kappa} && - \text{Increase $w$ linearly} \\
& T_\downarrow\textsubscript{linear}(i) = 1 - \tfrac{i}{\kappa} && - \text{Decrease $w$ linearly}   \\
& T\textsubscript{constant}(i) = 0.5 && - \text{Keep $w$ constant}   \\
& T\textsubscript{sigmoid}(i) = \sigma((i - \lambda)  * \kappa) && - \text{Increase $w$ non-linearly}
\end{alignat*} \\
\noindent with $\kappa$ and $\lambda$ being constants and $\sigma$ being the sigmoid function.
We illustrate these toy policies and their performance on CIFAR10 in Appendix~\ref{app:toy_teachers}. 
Interestingly, some of these toy curricula produce learning advantages akin to commentaries.
In fact, effective curricula shift weights from smaller towards larger values throughout training, suggesting that such shifts are underpinning the success of the curriculum.


\section{Curriculum-Adam interactions}
\label{sec:cl_adam_interaction}
We have seen how simple shifts from small to large loss weights can mimic the effects of the commentary curriculum.
How is this possible?
First, we know that the effect works across datasets, modalities and models and must therefore originate in the data- and model-agnostic optimisation process.
In our case, optimisation centres around the Adam optimiser \citep{kingma2014adam}.
Second, the effective component in our toy curricula is the \emph{change} of weighting with time. 
In the Adam optimiser, the only components sensitive to changes with time are the two momentum terms $m_i$ and $v_i$.
In the following, we will analyse the momentum terms of Adam (see Algorithm~\ref{adam_alg}) to find a potential source of the learning advantages in commentaries.

\begin{algorithm}
\caption{Adam (simplified)}
\label{adam_alg}
\begin{algorithmic}[1]
\STATE \textbf{Inputs}:  $\gamma$ (lr), $\beta_1$, $\beta_2$ (decay-rates), $\theta$ (parameters), $f(\theta)$ (objective)
\STATE \textbf{initialise} $m_i \gets 0$, $v_i \gets 0$
\FOR{$i \in\{1,\ldots I\}$}
        \STATE $g_i \gets \Delta_\theta f_i(\theta_{i-1})$
        
        \STATE $m_i \gets \beta_1 m_{i-1} + (1 - \beta_1)g_i$
		\STATE $v_i \gets \beta_2 v_{i-1} + (1 - \beta_2)g^2_i$
		\STATE $\hat{m_i} \gets m_{i} / (1 - \beta^i_1)$
		\STATE $\hat{v_i} \gets v_{i} / (1 - \beta^i_2)$
		\STATE $\Delta\theta_i \gets \hat{m_i}/(\sqrt{\hat{v_i}} + \epsilon)$
		\STATE $\theta_i \gets \theta_{i-1} - \gamma\Delta\theta_i$
   \ENDFOR
   \STATE \textbf{return} $\theta_i$

\end{algorithmic}
\end{algorithm}

\paragraph{Asymmetric momenta}
In the Adam algorithm, both momenta, $m_i$ and $v_i$, are determined by the current gradient $g_i$ as well as their previous states ($m_{i-1}$ and $v_{i-1}$, respectively). They are used to calculate the final parameter update $\Delta\theta_i$.
For either term, the influences of past states is decayed  at their own rate $\beta_1$ and $\beta_2$ (see line 5 \& line 6).
By default, $\beta_1$ and $\beta_2$ are set to largely different values\footnote{\citet{kingma2014adam} recommend: $\beta_1$ = 0.9; $\beta_2$ = 0.999}.
Its progressive decay rate $\beta_1$ makes $m_i$ more dependent on immediately preceding states, while $v_i$ is largely influenced by more distant states.
Both momenta are therefore asymmetric in their past dependence.
To calculate the parameter update $\Delta\theta_i$ (line 9), the faster decaying term $m_i$ is divided by the square root of the slower decaying $v_i$.
This step is done to normalise the size\footnote{For simplicity, we refer to the l2-norm (calculate as $|v|_2 = \sqrt[]{v^2_1 + .. + v^2_n}$) of a vector as its `size' throughout this and the following sections. Further, we simplify its notation to be $|v|$.} of the parameter-update $|\Delta\theta_i|$, and in a regular setup the asymmetry of decay is irrelevant as the size of $m_i$ and $v_i$ remains (more or less) constant throughout training. 

\paragraph{Interaction between momenta and curricula}
In our toy experiments (and in commentaries), we scale our losses (and therewith the gradients $g_i$) by $w_i$ to become larger with time.
If we systematically increase the size of $g_i$ with time, $|m_t|$ grows faster than $|v_t|$.
By normalising the $m_t$ term by the therewith smaller $v_t$ term, we artificially increase the size of the update $|\Delta\theta_i|$.
There thus exists an interaction between the momentum terms and the shape of the curriculum.
This effect is easy to empirically exemplify in a minimal example.

\begin{figure}
  \centering
  \begin{subfigure}{0.49\linewidth}
    \includegraphics[width=\linewidth]{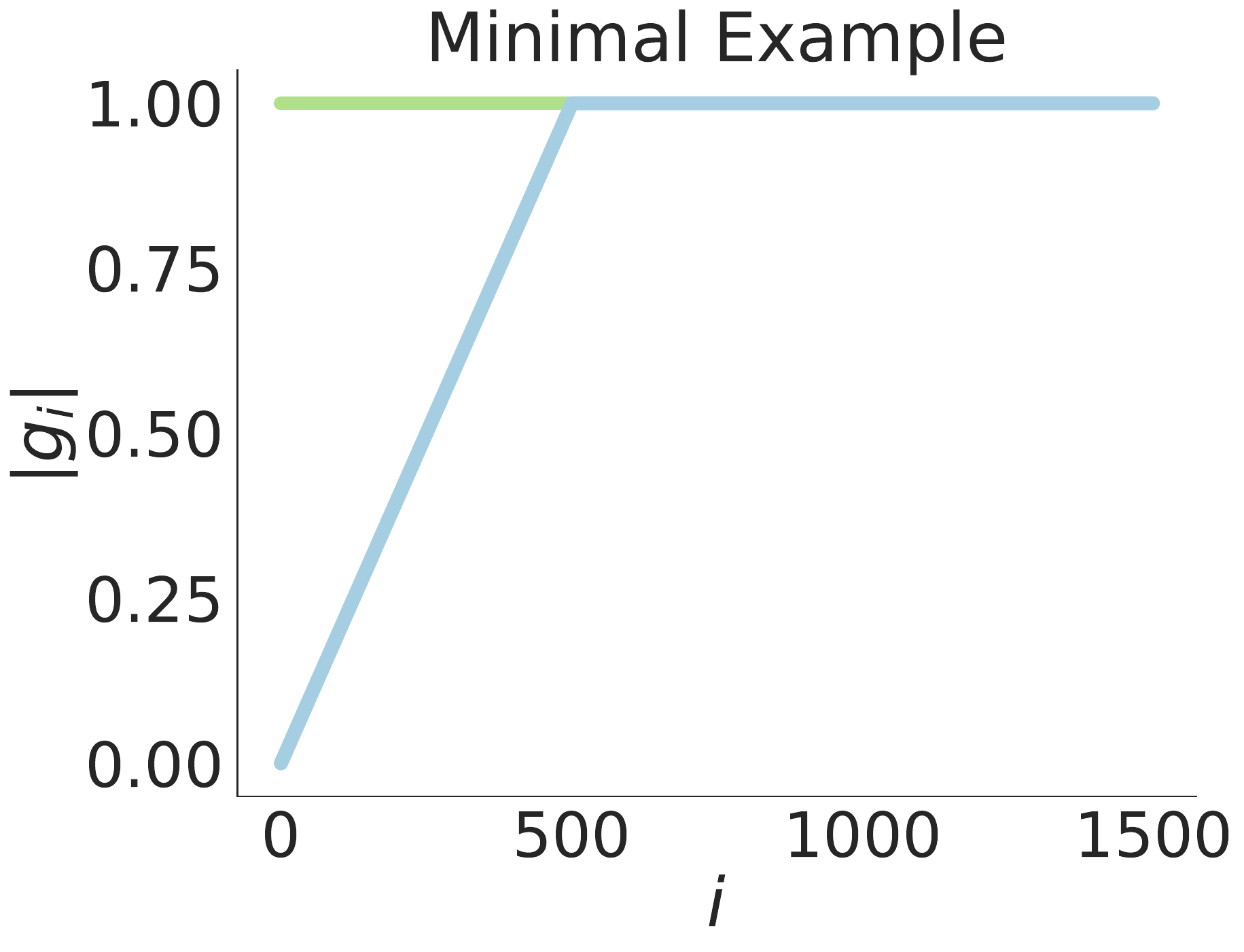}
    \caption{}
    \label{subfig:toy_grads}
  \end{subfigure}
  \begin{subfigure}{0.49\linewidth}
    \includegraphics[width=\linewidth]{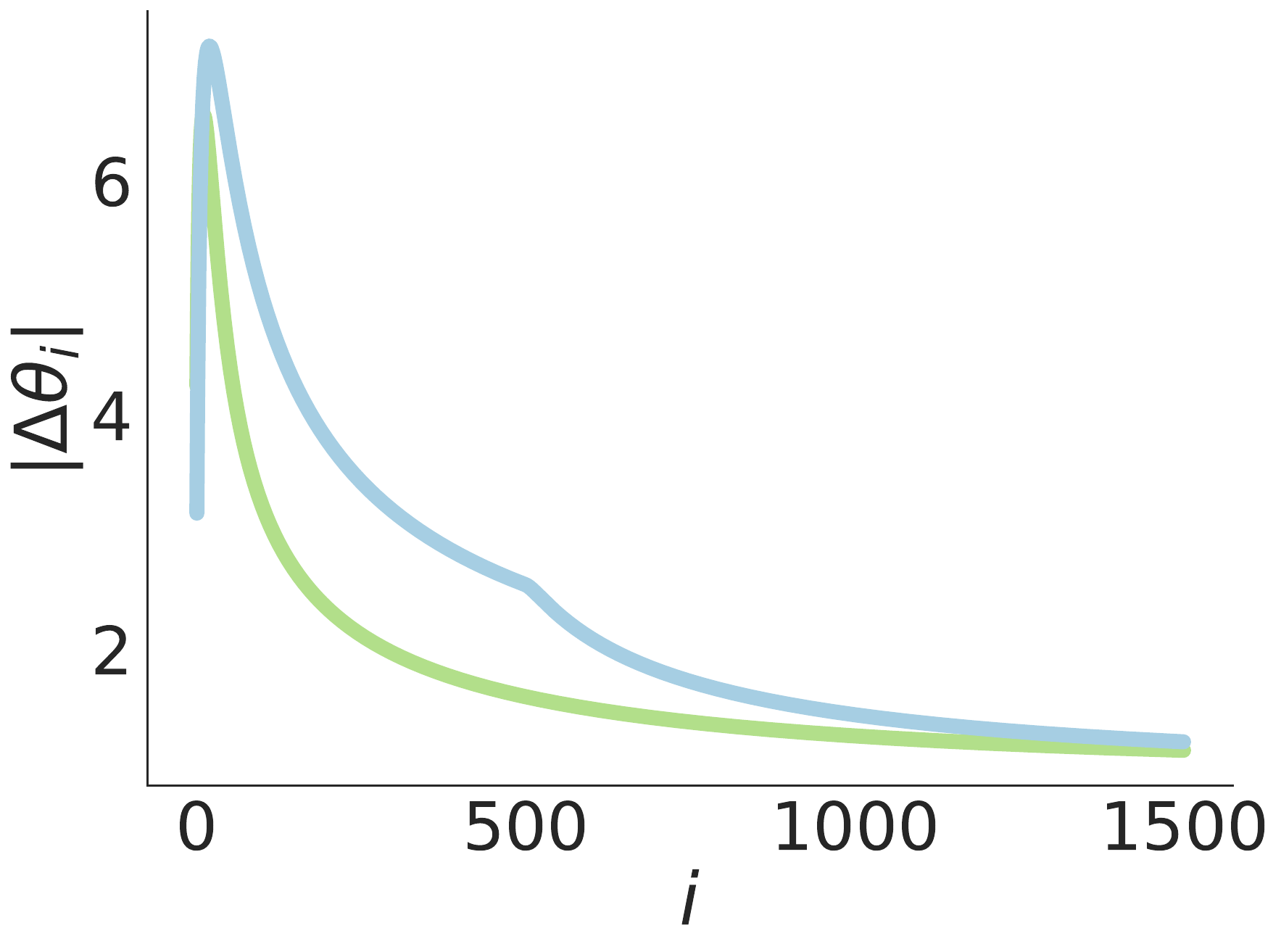} 
    \caption{}
    \label{subfig:toy_deltas}
  \end{subfigure}
  
  \caption{Minimal example with a single parameter: If we increase the gradient of the parameter linearly (left), Adam produces larger parameter updates $|\Delta\theta_i|$ compared to a constant gradient size (right).
 }
  \label{fig:minimal_example}
\end{figure}

We consider a simple case with only a single parameter.
We create two conditions: In the first condition, we linearly increase the gradient size $|g_i|$ from 0 to 1 where it levels off (similar to the linear toy curriculum).
In the baseline condition, the $|g_i|$ remains fixed at the value of 1 (Figure~\ref{subfig:toy_grads}).
For the first condition, the size of the update returned by Adam is systematically larger compared to the baseline condition (Figure~\ref{subfig:toy_deltas}).
We hypothesise that this scaling of $|\Delta\theta_i|$ is behind the observed learning improvements of commentaries and our toy experiments.
We can test whether this is true by checking the following two entailments:

\begin{figure*}
  \centering
  \begin{subfigure}{0.30\linewidth}
    \includegraphics[width=\linewidth]{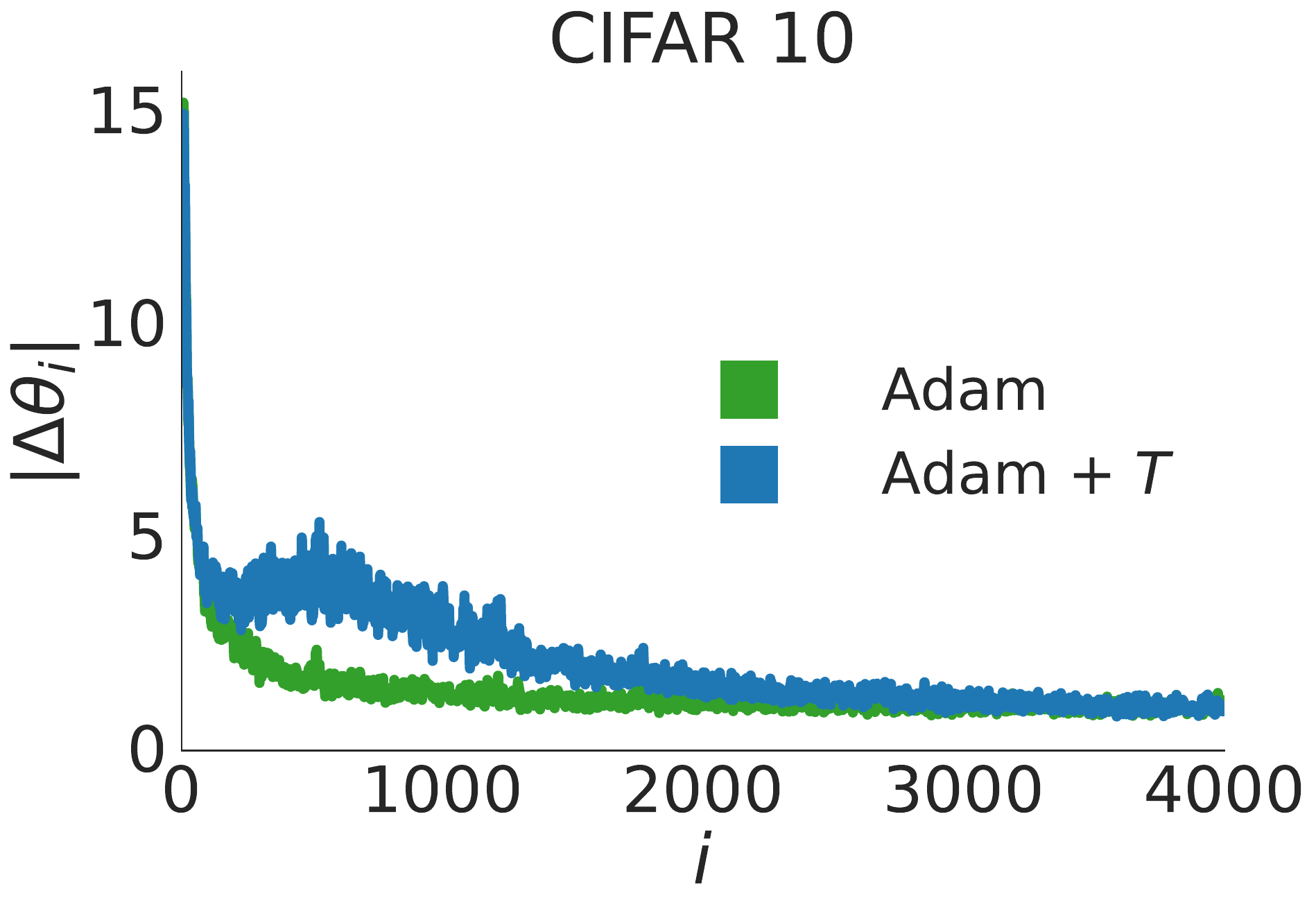}
    \caption{}
    \label{subfig:commentaries_deltas}
  \end{subfigure}
  \begin{subfigure}{0.30\linewidth}
    \includegraphics[width=\linewidth]{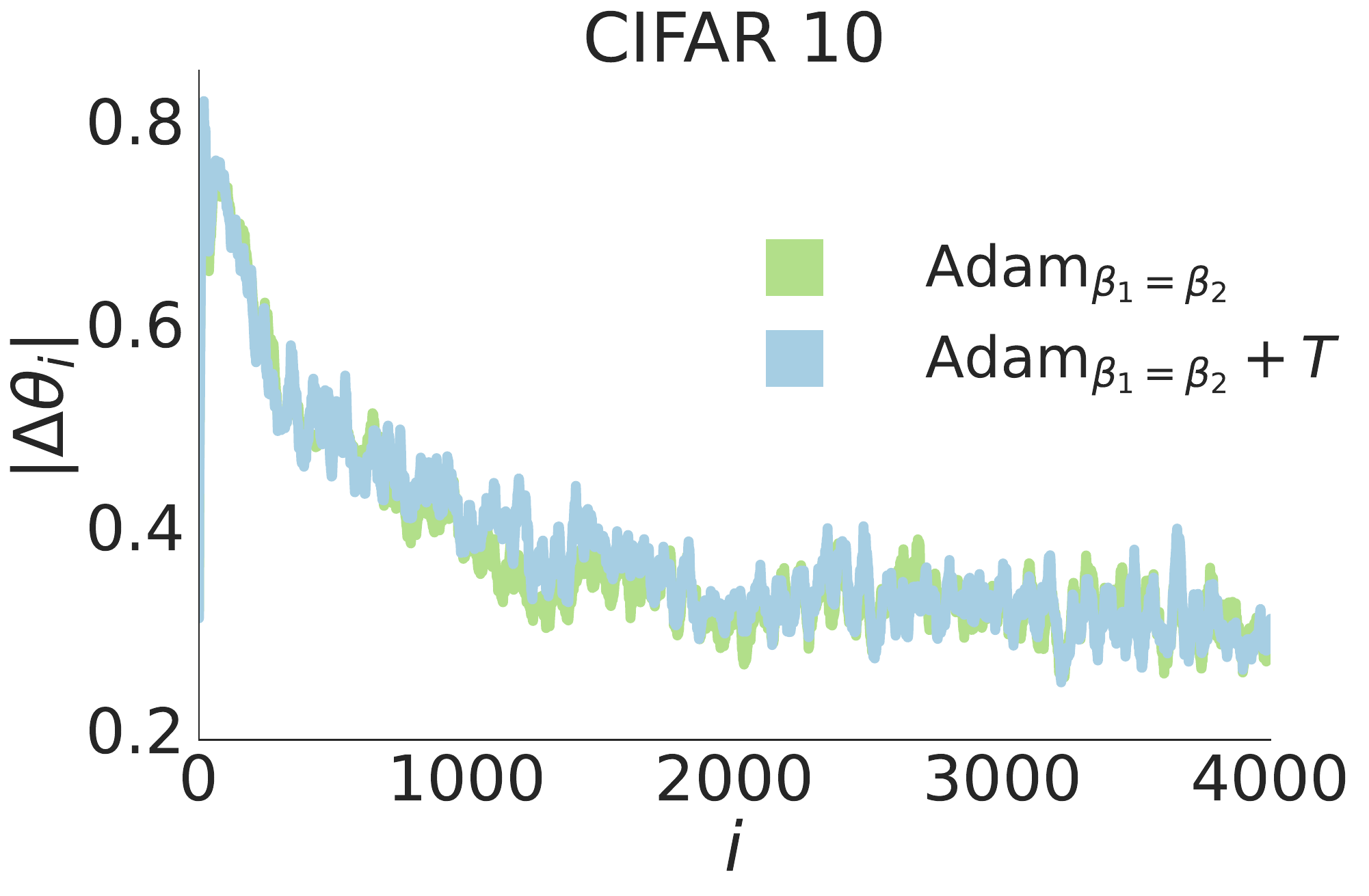}
    \caption{}
    \label{subfig:beta_equal_norms}
  \end{subfigure}
  \begin{subfigure}{0.30\linewidth}
    \includegraphics[width=\linewidth]{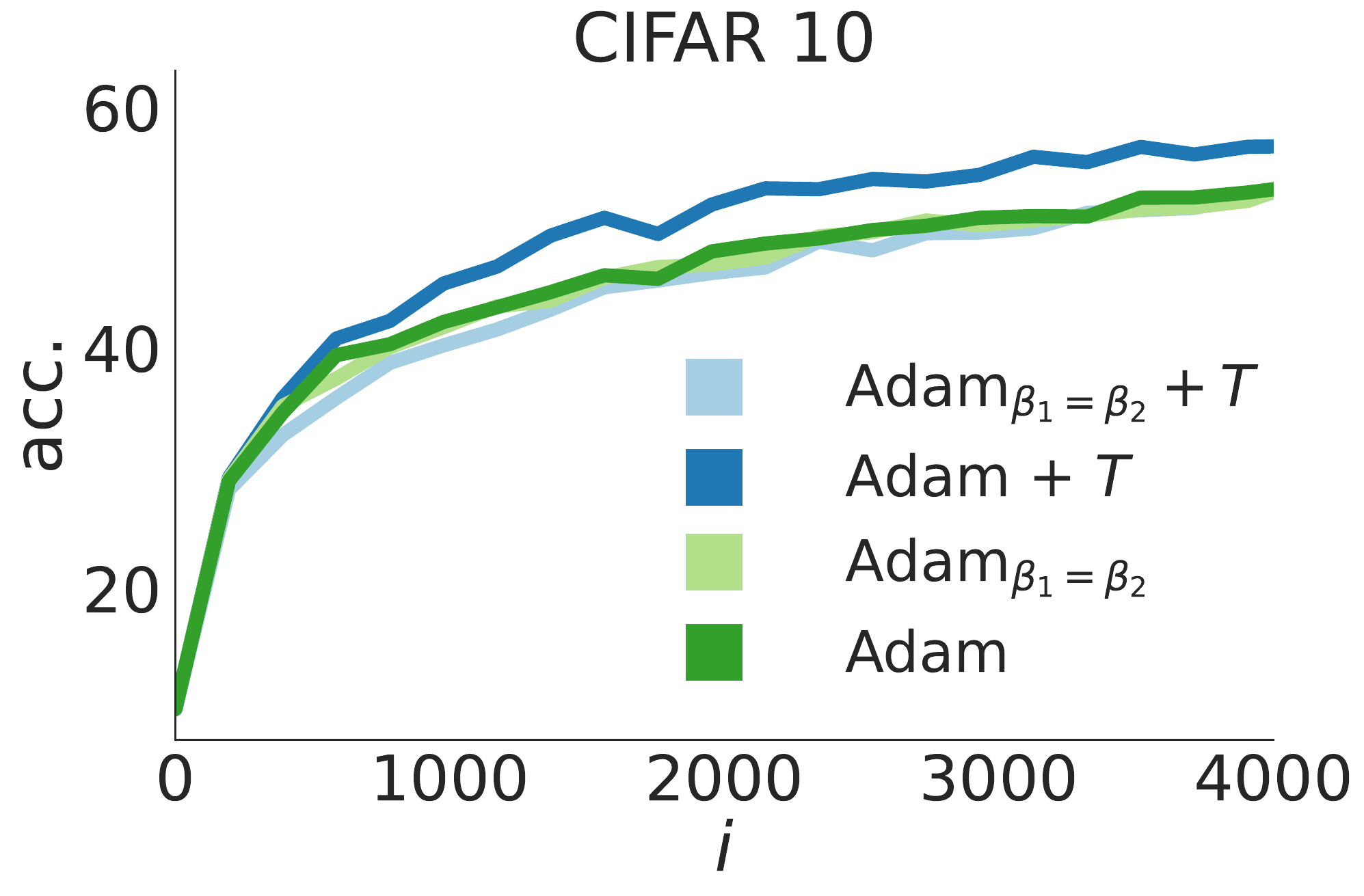} 
    \caption{}
    \label{subfig:beta_equal_performance}
  \end{subfigure}
  
  \caption{(a) Akin to the minimal example in Figure~\ref{fig:minimal_example}, the commentary teacher also produces larger parameter update $|\Delta\theta_i|$ due to Curriculum-Adam-interactions. (b) We can neutralise the Curriculum-Adam-interactions by setting Adam's $\beta$ parameters to equal values ($\beta_1$ = $\beta_2$ = 0.99).
  With this intervention, the difference of $|\Delta\theta_i|$ that we observed in (a) vanishes. As a consequence, the performance of the commentaries' curriculum drops to the baseline (c). This shows how the interaction-dependent increase in $|\Delta\theta_i|$ is crucial for the learning speed gains of commentaries. 
 }
  \label{fig:beta_equal}
\end{figure*}

\begin{description}
\item{\textbf{Entailment 1}: } The size of the update $|\Delta\theta_i|$ for commentaries is larger than for the baseline while $w_i$ increases in size. Afterwards, $|\Delta\theta_i|$ drops to normal levels.
	\item{\textbf{Entailment 2}: } Making $m_i$ and $v_i$ equally dependent on past $|g|$ by setting the decay-factors to $\beta_1$ = $\beta_2$ leads to the curriculum losing its effect.
\end{description}

We go on to empirically test these entailments for commentaries.
Moreover, other curricula that cause systematic shifts in gradients sizes can result in similar effects.
We, therefore, continue to also test different other curricula.

\subsection{Interactions with Commentaries}
\label{subsec:CL_Adam_commentaries}

\paragraph{Experiments}
\label{para:experiments_comms}
For the first set of experiments, we apply minimal necessary changes to the original setup of \citet{raghu2020teaching}.
We reutilise the teacher model from \ref{subsec:comm_exp_setup} to train a new target model on the CIFAR10-dataset \citep{krizhevsky2009learning}. %

We test the first entailment by recording the size of the student's parameter updates $|\Delta\theta_i|$ and of the baseline model without loss reweighting during the training.
Comparing the two, we find that the model with loss-reweighing experiences an increase in $|\Delta\theta_i|$ compared to training without a teacher (Figure~\ref{subfig:commentaries_deltas}). 
The `boost' in the update norm corresponds neatly to the range of iterations $i$ in which $w_i$ increases starkly (compare Figure~\ref{subfig:weight_stats_commentaries}). 
Our observations are very similar to the minimal example described in Section~\ref{sec:cl_adam_interaction} and are in line with \textbf{Entailment 1}. %
This experiment provides supportive evidence for our hypothesis, but it is not yet sufficient: the observed `boost' could potentially arise from factors such as the enhanced properties of the optimization landscape, as discussed in \citet[][]{bengio2009curriculum}.

We rule out such alternative explanations by eliminating the effect of the \emph{Adam-curriculum}-interactions while keeping potential other effects of the curriculum unaffected.
To do so, we equalise the past dependence of the momentum terms by setting both of Adam's $\beta$s to the same value ($\beta_1$ = $\beta_2$ = $0.99$). 
This results in Adam becoming equivalent to standard stochastic gradient descent (SGD) with a normalised momentum term\footnote{The $\beta$s can be chosen in the same way as the decay factor $\beta$ in SGD}.

We train an additional set of target models with this alternative hyperparameter setting.
As a consequence, the difference in $|\Delta\theta_i|$ disappears (see Figure~\ref{subfig:beta_equal_norms}) and the learning advantage in accuracy vanishes (Figure~\ref{subfig:beta_equal_performance}).
This verifies \textbf{Entailment 2}.

We have seen so far that the \emph{Adam-curriculum}-interactions scale the parameter updates $|\Delta\theta|$. 
Doing so should ultimately have the same effect as increasing the learning-rate $\gamma$ (see line 10 in Algorithm~\ref{adam_alg}).
Hence, instead of using a curriculum, we can simply adjust $\gamma$.
We show that this has the same effect by training three sets of target models (with and without loss-reweighting) with learning rates spanning three orders of magnitude. 
We find that only for very low values of $\gamma$ the compensating effect of commentaries helps learning (Figure~\ref{fig:CIFAR_higher_lr} in Appendix~\ref{app:replication}). 
With a properly tuned $\gamma$, the difference between the baseline and commentary condition vanishes.



\paragraph{Conclusions}
We can summarise the results of our first set of experiments as follows: 
First, the effectiveness of the commentaries curriculum is a result of \emph{Adam-curriculum}-interactions that scale parameter updates to become larger.
Second,  we can eliminate the effect of interactions by setting Adam's $\beta$-parameters to equal values. 
This eliminates any learning advantage.
Third, the observed learning advantages are only possible due to suboptimal hyperparameters; as soon as we set hyperparameters optimally, vanilla Adam outperforms the curriculum.

Parametric approaches to curriculum learning are especially vulnerable to this interaction, as they can adapt their schedule function to optimally compensate for suboptimal hyperparameters. 
However, we expect that also hand-crafted curricula can be affected by this interaction.
In what follows, we investigate the impact of Curriculum-Adam-interactions on other types of curricula.

\begin{figure*}
  \centering
  \begin{subfigure}{0.39\linewidth}
    \includegraphics[width=\linewidth]{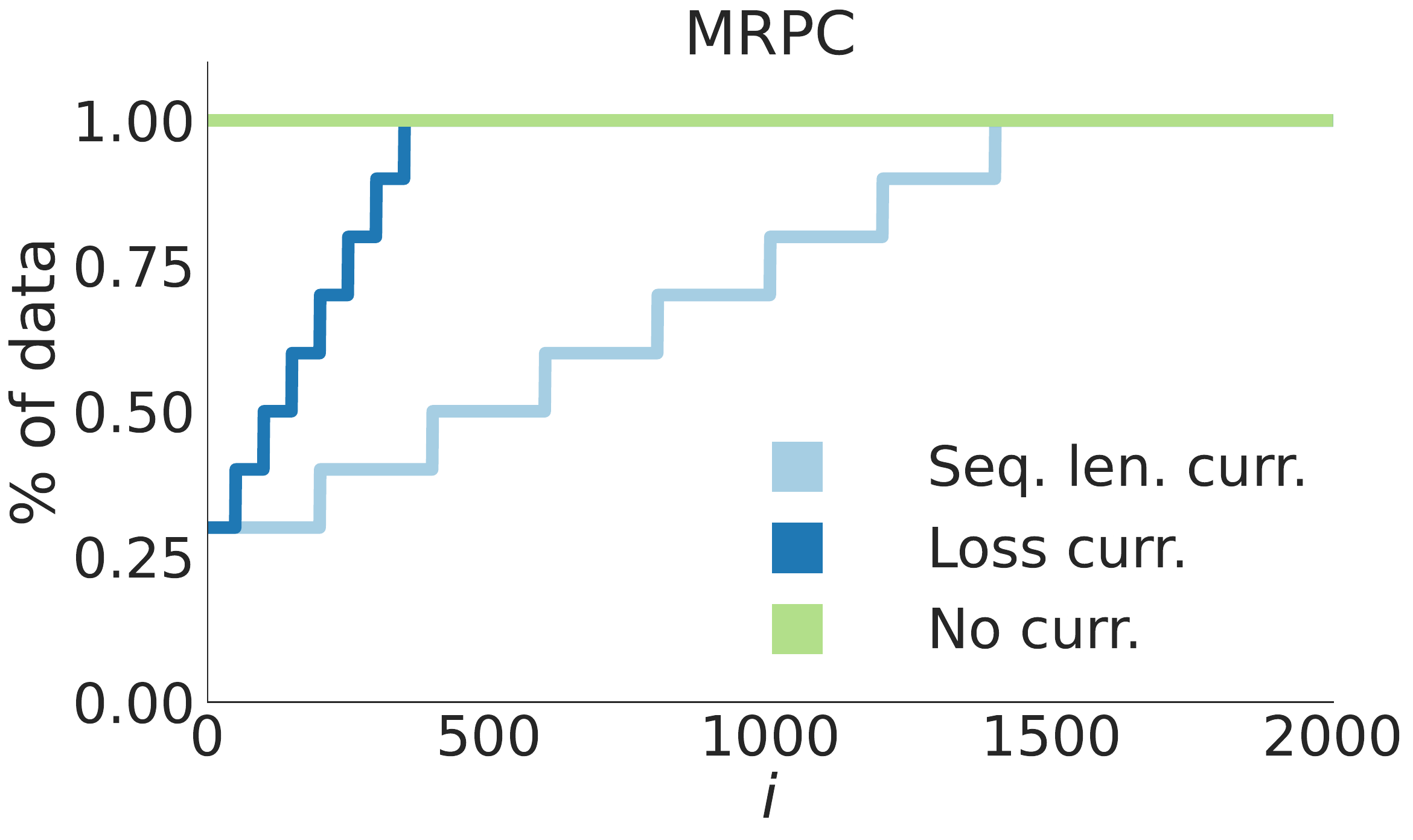}
    \caption{}
    \label{subfig:schedule_functions}
  \end{subfigure}
  \begin{subfigure}{0.37\linewidth}
    \includegraphics[width=\linewidth]{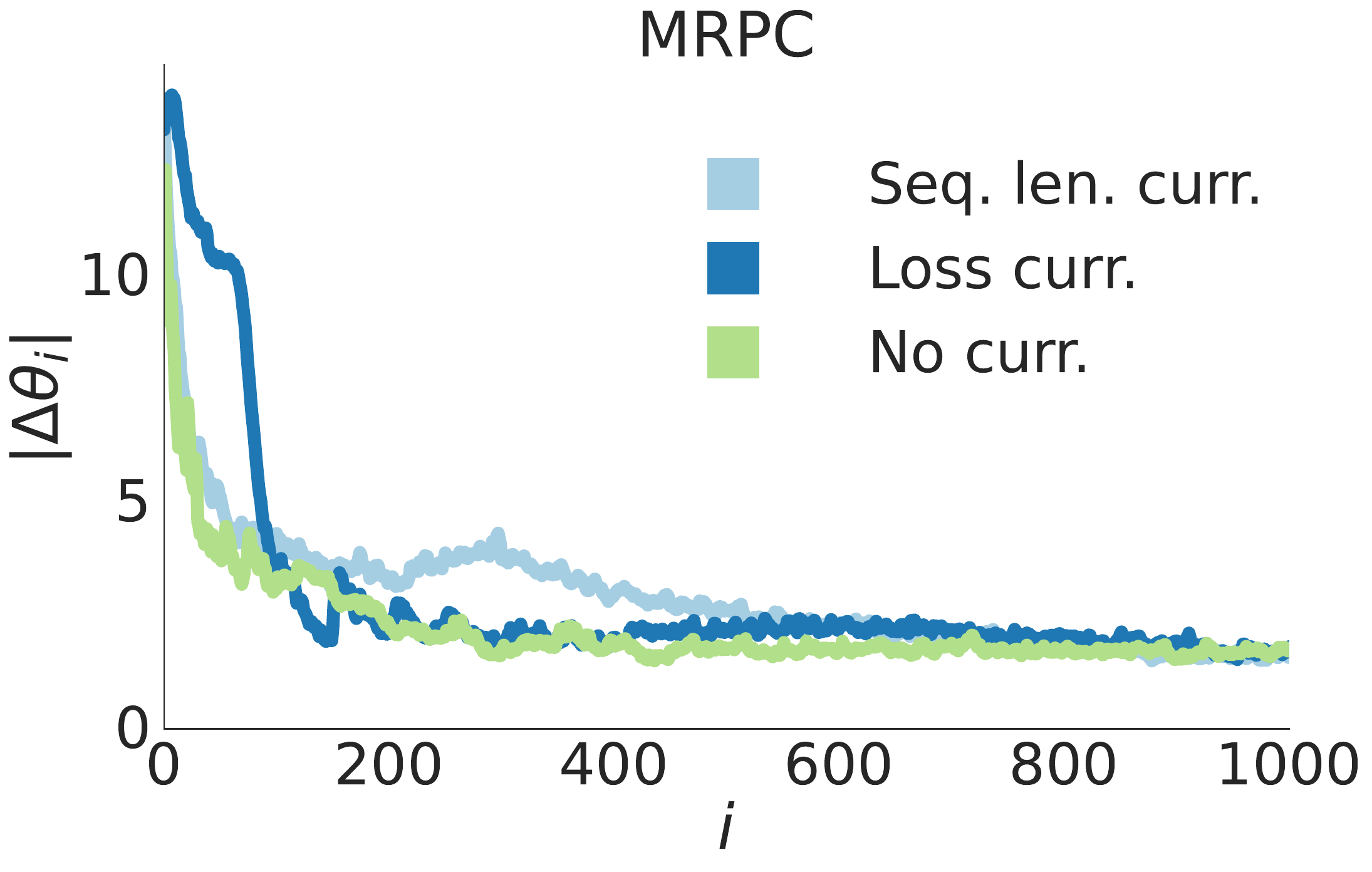} 
    \caption{}
    \label{subfig:deltas_mrpc_handcrafted}
  \end{subfigure}
  
  \caption{On the left we illustrate the schedule functions used for our experiments in Section~\ref{subsec:hc_curricula}. On the right side, we see the corresponding sizes of parameter updates $|\Delta\theta|$. We see an increase in parameter updates at the largest relative change of the data distribution.} %
  \label{fig:hand_crafted_curricula_schedule_function_and_delta_thetas}
\end{figure*}

\subsection{Interactions with hand-crafted curricula}
\label{subsec:hc_curricula}

We now investigate other common hand-crafted and non-parametric curricula, such as pacing via sentence length or loss \citep[e.g.][]{spitkovsky2009baby, platanios2019competence, tay2019simple}.
These curricula do not have explicit shifts of gradient sizes from small to large. 
However, we have reason to believe that they might be affected by interactions with Adam nevertheless:
We expect that difficulty measures like \textbf{sequence lengths} \citep{spitkovsky2009baby, platanios2019competence, tay2019simple} or \textbf{loss} \citep{kumar2010self} are oftentimes correlated with the size of the gradients $|g|$ that they produce. 
We find that this is the case when we finetune RoBERTa\textsubscript{BASE} \citep{liu2019roberta} on a selection of GLUE-tasks (a plot relating sequence lengths and losses to the size of the resulting gradients $|g|$ can be found in Appendix~\ref{app:correlations_diffm_and_gradients}). 

A curriculum that orders training examples according to these difficulty measures, hence, also implicitly orders them according to their gradient sizes.
As a consequence, classical hand-crafted curricula potentially also trigger interactions with Adam.
We will test such curricula for interactions in the following paragraph.

\paragraph{Experiments}
We implement two simple but common hand-crafted curriculum setups which use sequence length (1) and cross-entropy-loss (2) as difficulty measures and employ the discrete schedule functions shown in Figure~\ref{subfig:schedule_functions}.
Ahead of training, we order the training data according to either their sequence length or the losses obtained by a RoBERTa\textsubscript{BASE} model that we finetuned on the respective task.
The curriculum randomly samples from an incrementally larger portion of the ordered dataset.
We determined the hyperparameters of the schedule functions by conducting grid-search, determining the best-performing setup on a subset of the validation data.
We then finetune RoBERTa\textsubscript{BASE} \citep{liu2019roberta} with both, an optimal and a slightly suboptimal learning rate, on the MRPC-task from the GLUE-dataset \citep{wang2018glue}.

Table~\ref{tab:hc_curriculum_mrpc} reports results for the hand-crafted curricula. %
If the learning rate is low, both of our improvised curricula let RoBERTa learn much faster compared to training without curriculum (as shown by the performance after $i$ = 750 steps).
However, as soon as we increase the learning rate to an optimal level vanilla Adam outperforms all other conditions. 
Analogously to our experiments with commentaries, we find the size of the parameter updates $|\Delta\theta|$ to be increased during the time of the largest change in data distribution (Figure~\ref{subfig:deltas_mrpc_handcrafted}).
The gain in $|\Delta\theta|$ for hand-crafted curricula is not as prolonged as for commentaries. 
This makes sense, as the shift in training distribution is especially large at the beginning of training, while in later steps the relative change is neglectable.
Despite gains in $|\Delta\theta|$ being relatively small and early in training, we observe that they are crucial for the performance gains of the curricula: If we eliminate the interaction with Adam by setting $\beta_1 = \beta_2$ the advantage of this simple curriculum vanishes (see Figure~\ref{subfig:predefined_curricula_perf_beta_equal} in Appendix~\ref{app:learning_curves_hc}). 

\begin{table}
\begin{center}
\begin{small}
\begin{sc}
\begin{tabular}{llccr}
\toprule
Setup & & $i$ = 750 & converged  \\
\midrule
 & No Curr.  & 77.8$\pm$1.5  & 88.2$\pm$0.6    \\
$\gamma$\textsubscript{LOW} + & Seq. len. Curr   & 82.8$\pm$1.1  & 87.8$\pm$0.5   \\
 & Loss Curr   & \textbf{84.2$\pm$0.51}  & 88.4$\pm$0.7    \\
\midrule
 & No Curr.   & \textbf{87.6$\pm$1.4} &  90.1$\pm$0.3  \\
$\gamma$\textsubscript{OPTIMAL} + & Seq. len. Curr   & 83.2 $\pm$3.5 & 89.1$\pm$1.4  \\
 & Loss Curr   & 79.5$\pm$9.6 & 90.0$\pm$0.9     
\\
  
\bottomrule
\end{tabular}
\caption{MRPC-validation accuracies of RoBERTa\textsubscript{BASE} for hand-crafted curricula at an early stage ($i$ = 750) and after convergence.}
\label{tab:hc_curriculum_mrpc}
\end{sc}
\end{small}
\end{center}
\end{table}

\paragraph{Conclusions}
In summary, we find that interactions between curriculum structure and Adam can also occur in hand-crafted curricula. This is the case if the difficulty measures are correlated with the gradient norms that they produce (e.g. if long sequences produce small gradients and short sequences produce large gradients).
The interaction produces learning speed improvements when finetuning RoBERTa\textsubscript{BASE} with slightly suboptimal learning rates and, again, Adam with optimal hyperparameter settings is able to outperform the curriculum.


\section{Practical implications and general conclusion}
\label{sec:conclusion}

In this paper, we show how optimising a model using a curriculum in combination with Adam can lead to unintended interactions between the two.
These interactions scale the parameter updates applied to the model, equivalent to a temporary scaling of the learning rate. 
Larger parameter updates lead to faster learning when hyperparameters (such as the learning rate) are chosen suboptimally \citep[as shown for][and exemplary for common hand-crafted curricula]{raghu2020teaching}. 
However, if hyperparameters are chosen correctly, vanilla Adam without curriculum always outperforms any curriculum learning approach that we employed.
Our analysis contains a large range of settings, including different training regimes (toy-setting, training from scratch and fine-tuning pretrained models), different modalities (vision and language) and different types of curricula (automated vs. hand-crafted)'.

We show that non-functional curricula can be remarkably deceptive:
the commentaries curriculum closely resembles known curricula from the literature, even though it ultimately works for very different reasons.
Our results warrant special caution for future research: research in curriculum learning using Adam has to be accompanied by a rigorous hyperparameter search to make reliable claims about the success of the curriculum beyond reducing the need for hyperparameter selection.

\section*{Acknowledgments}
...
\bibliography{additional_refs.bib}

\begin{thebibliography}{33}
\providecommand{\natexlab}[1]{#1}

\bibitem[{Allgower and Georg(1980)}]{allgower1980numerical}
Allgower, E.~L.; and Georg, K. 1980.
\newblock \emph{Numerical continuation methods: an introduction}, volume~13.
\newblock Springer Science \& Business Media.

\bibitem[{Bengio et~al.(2009)Bengio, Louradour, Collobert, and
  Weston}]{bengio2009curriculum}
Bengio, Y.; Louradour, J.; Collobert, R.; and Weston, J. 2009.
\newblock Curriculum learning.
\newblock In \emph{Proceedings of the 26th annual international conference on
  machine learning}, 41--48.

\bibitem[{Campos(2021)}]{campos2021curriculum}
Campos, D. 2021.
\newblock Curriculum learning for language modeling.
\newblock \emph{arXiv preprint arXiv:2108.02170}.

\bibitem[{Dolan and Brockett(2005)}]{dolan2005automatically}
Dolan, B.; and Brockett, C. 2005.
\newblock Automatically constructing a corpus of sentential paraphrases.
\newblock In \emph{Third International Workshop on Paraphrasing (IWP2005)}.

\bibitem[{Elman(1993)}]{elman1993learning}
Elman, J.~L. 1993.
\newblock Learning and development in neural networks: The importance of
  starting small.
\newblock \emph{Cognition}, 48(1): 71--99.

\bibitem[{Fan et~al.(2018)Fan, Tian, Qin, Li, and Liu}]{fan2018learning}
Fan, Y.; Tian, F.; Qin, T.; Li, X.-Y.; and Liu, T.-Y. 2018.
\newblock Learning to Teach.
\newblock In \emph{International Conference on Learning Representations}.

\bibitem[{Hacohen and Weinshall(2019)}]{hacohen2019power}
Hacohen, G.; and Weinshall, D. 2019.
\newblock On the power of curriculum learning in training deep networks.
\newblock In \emph{International Conference on Machine Learning}, 2535--2544.
  PMLR.

\bibitem[{He et~al.(2016)He, Zhang, Ren, and Sun}]{he2016deep}
He, K.; Zhang, X.; Ren, S.; and Sun, J. 2016.
\newblock Deep residual learning for image recognition.
\newblock In \emph{Proceedings of the IEEE conference on computer vision and
  pattern recognition}, 770--778.

\bibitem[{Jiang et~al.(2018)Jiang, Zhou, Leung, Li, and
  Fei-Fei}]{jiang2018mentornet}
Jiang, L.; Zhou, Z.; Leung, T.; Li, L.-J.; and Fei-Fei, L. 2018.
\newblock Mentornet: Learning data-driven curriculum for very deep neural
  networks on corrupted labels.
\newblock In \emph{International conference on machine learning}, 2304--2313.
  PMLR.

\bibitem[{Kim and Choi(2018)}]{kim2018screenernet}
Kim, T.-H.; and Choi, J. 2018.
\newblock Screenernet: Learning self-paced curriculum for deep neural networks.
\newblock \emph{arXiv preprint arXiv:1801.00904}.

\bibitem[{Kingma and Ba(2014)}]{kingma2014adam}
Kingma, D.~P.; and Ba, J. 2014.
\newblock Adam: A method for stochastic optimization.
\newblock \emph{arXiv preprint arXiv:1412.6980}.

\bibitem[{Kocmi and Bojar(2017)}]{kocmi2017curriculum}
Kocmi, T.; and Bojar, O. 2017.
\newblock Curriculum Learning and Minibatch Bucketing in Neural Machine
  Translation.
\newblock In \emph{Proceedings of the International Conference Recent Advances
  in Natural Language Processing, RANLP 2017}, 379--386.

\bibitem[{Krizhevsky, Hinton et~al.(2009)}]{krizhevsky2009learning}
Krizhevsky, A.; Hinton, G.; et~al. 2009.
\newblock Learning multiple layers of features from tiny images.

\bibitem[{Krueger and Dayan(2009)}]{krueger2009flexible}
Krueger, K.~A.; and Dayan, P. 2009.
\newblock Flexible shaping: How learning in small steps helps.
\newblock \emph{Cognition}, 110(3): 380--394.

\bibitem[{Kumar, Packer, and Koller(2010)}]{kumar2010self}
Kumar, M.; Packer, B.; and Koller, D. 2010.
\newblock Self-paced learning for latent variable models.
\newblock \emph{Advances in neural information processing systems}, 23.

\bibitem[{Liu et~al.(2018)Liu, He, Liu, Zhao et~al.}]{liu2018curriculum}
Liu, C.; He, S.; Liu, K.; Zhao, J.; et~al. 2018.
\newblock Curriculum Learning for Natural Answer Generation.
\newblock In \emph{IJCAI}, 4223--4229.

\bibitem[{Liu et~al.(2019)Liu, Ott, Goyal, Du, Joshi, Chen, Levy, Lewis,
  Zettlemoyer, and Stoyanov}]{liu2019roberta}
Liu, Y.; Ott, M.; Goyal, N.; Du, J.; Joshi, M.; Chen, D.; Levy, O.; Lewis, M.;
  Zettlemoyer, L.; and Stoyanov, V. 2019.
\newblock Roberta: A robustly optimized bert pretraining approach.
\newblock \emph{arXiv preprint arXiv:1907.11692}.

\bibitem[{Mart{\'\i}nez~Alonso et~al.(2017)Mart{\'\i}nez~Alonso, Agi{\'c},
  Plank, and S{\o}gaard}]{alonso2017parsing}
Mart{\'\i}nez~Alonso, H.; Agi{\'c}, {\v{Z}}.; Plank, B.; and S{\o}gaard, A.
  2017.
\newblock Parsing {U}niversal {D}ependencies without training.
\newblock In \emph{Proceedings of the 15th Conference of the {E}uropean Chapter
  of the Association for Computational Linguistics: Volume 1, Long Papers},
  230--240. Valencia, Spain: Association for Computational Linguistics.

\bibitem[{Narvekar et~al.(2020)Narvekar, Peng, Leonetti, Sinapov, Taylor, and
  Stone}]{narvekar2020curriculum}
Narvekar, S.; Peng, B.; Leonetti, M.; Sinapov, J.; Taylor, M.~E.; and Stone, P.
  2020.
\newblock Curriculum learning for reinforcement learning domains: A framework
  and survey.
\newblock \emph{The Journal of Machine Learning Research}, 21(1): 7382--7431.

\bibitem[{Ott et~al.(2019)Ott, Edunov, Baevski, Fan, Gross, Ng, Grangier, and
  Auli}]{ott2019fairseq}
Ott, M.; Edunov, S.; Baevski, A.; Fan, A.; Gross, S.; Ng, N.; Grangier, D.; and
  Auli, M. 2019.
\newblock fairseq: A Fast, Extensible Toolkit for Sequence Modeling.
\newblock In \emph{Proceedings of the 2019 Conference of the North American
  Chapter of the Association for Computational Linguistics (Demonstrations)},
  48--53.

\bibitem[{Penha and Hauff(2020)}]{penha2020curriculum}
Penha, G.; and Hauff, C. 2020.
\newblock Curriculum learning strategies for ir.
\newblock In \emph{European Conference on Information Retrieval}, 699--713.
  Springer.

\bibitem[{Platanios et~al.(2019)Platanios, Stretcu, Neubig, P{\'o}czos, and
  Mitchell}]{platanios2019competence}
Platanios, E.~A.; Stretcu, O.; Neubig, G.; P{\'o}czos, B.; and Mitchell, T.
  2019.
\newblock Competence-based Curriculum Learning for Neural Machine Translation.
\newblock In \emph{Proceedings of the 2019 Conference of the North American
  Chapter of the Association for Computational Linguistics: Human Language
  Technologies, Volume 1 (Long and Short Papers)}, 1162--1172.

\bibitem[{Raghu et~al.(2020)Raghu, Raghu, Kornblith, Duvenaud, and
  Hinton}]{raghu2020teaching}
Raghu, A.; Raghu, M.; Kornblith, S.; Duvenaud, D.; and Hinton, G. 2020.
\newblock Teaching with Commentaries.
\newblock In \emph{International Conference on Learning Representations}.

\bibitem[{Rohde and Plaut(1999)}]{rohde1999language}
Rohde, D.~L.; and Plaut, D.~C. 1999.
\newblock Language acquisition in the absence of explicit negative evidence:
  How important is starting small?
\newblock \emph{Cognition}, 72(1): 67--109.

\bibitem[{Soviany et~al.(2022)Soviany, Ionescu, Rota, and
  Sebe}]{soviany2022curriculum}
Soviany, P.; Ionescu, R.~T.; Rota, P.; and Sebe, N. 2022.
\newblock Curriculum learning: A survey.
\newblock \emph{International Journal of Computer Vision}, 1--40.

\bibitem[{Spitkovsky, Alshawi, and Jurafsky(2009)}]{spitkovsky2009baby}
Spitkovsky, V.~I.; Alshawi, H.; and Jurafsky, D. 2009.
\newblock Baby Steps: How “Less is More” in unsupervised dependency
  parsing.

\bibitem[{Spitkovsky, Alshawi, and Jurafsky(2010)}]{spitkovsky2010baby}
Spitkovsky, V.~I.; Alshawi, H.; and Jurafsky, D. 2010.
\newblock From baby steps to leapfrog: How “less is more” in unsupervised
  dependency parsing.
\newblock In \emph{Human Language Technologies: The 2010 Annual Conference of
  the North American Chapter of the Association for Computational Linguistics},
  751--759.

\bibitem[{Surkov, Mosin, and Yamshchikov(2022)}]{surkov2021data}
Surkov, M.; Mosin, V.; and Yamshchikov, I. 2022.
\newblock Do Data-based Curricula Work?
\newblock In \emph{Proceedings of the Third Workshop on Insights from Negative
  Results in NLP}, 119--128.

\bibitem[{Tay et~al.(2019)Tay, Wang, Luu, Fu, Phan, Yuan, Rao, Hui, and
  Zhang}]{tay2019simple}
Tay, Y.; Wang, S.; Luu, A.~T.; Fu, J.; Phan, M.~C.; Yuan, X.; Rao, J.; Hui,
  S.~C.; and Zhang, A. 2019.
\newblock Simple and Effective Curriculum Pointer-Generator Networks for
  Reading Comprehension over Long Narratives.
\newblock In \emph{Proceedings of the 57th Annual Meeting of the Association
  for Computational Linguistics}, 4922--4931.

\bibitem[{Vaswani et~al.(2017)Vaswani, Shazeer, Parmar, Uszkoreit, Jones,
  Gomez, Kaiser, and Polosukhin}]{vaswani2017attention}
Vaswani, A.; Shazeer, N.; Parmar, N.; Uszkoreit, J.; Jones, L.; Gomez, A.~N.;
  Kaiser, {\L}.; and Polosukhin, I. 2017.
\newblock Attention is all you need.
\newblock \emph{Advances in neural information processing systems}, 30.

\bibitem[{Wang et~al.(2018)Wang, Singh, Michael, Hill, Levy, and
  Bowman}]{wang2018glue}
Wang, A.; Singh, A.; Michael, J.; Hill, F.; Levy, O.; and Bowman, S.~R. 2018.
\newblock GLUE: A Multi-Task Benchmark and Analysis Platform for Natural
  Language Understanding.
\newblock In \emph{International Conference on Learning Representations}.

\bibitem[{Wang, Chen, and Zhu(2021)}]{wang2021survey}
Wang, X.; Chen, Y.; and Zhu, W. 2021.
\newblock A survey on curriculum learning.
\newblock \emph{IEEE Transactions on Pattern Analysis and Machine
  Intelligence}.

\bibitem[{Xu et~al.(2020)Xu, Zhang, Mao, Wang, Xie, and
  Zhang}]{xu2020curriculum}
Xu, B.; Zhang, L.; Mao, Z.; Wang, Q.; Xie, H.; and Zhang, Y. 2020.
\newblock Curriculum learning for natural language understanding.
\newblock In \emph{Proceedings of the 58th Annual Meeting of the Association
  for Computational Linguistics}, 6095--6104.

\end{thebibliography}

\newpage 
\onecolumn
\appendix
\section*{Supplementary Material}
\label{app:introduction_appendix}
The supplementary material contains additional information about exact hyperparameter settings for all experiments (Appendix~\ref{app:hyperparams}), additional learning curves for all replications of \citet{raghu2020teaching}'s experiments (Appendix~\ref{app:replication}) and results for our extension to GLUE-data (Appendix~\ref{app:extension_NLP}). 
Further, we illustrate the weighting policies of the toy-teachers from Section~\ref{subsec:data_independence} (Appendix~\ref{app:toy_teachers}), give empirical proof for relation of difficulty measures with their associated gradient norms $|g|$ (Appendix~\ref{app:correlations_diffm_and_gradients}) and provide the learning-curves of our finetuning experiments using hand-crafted curricula in Section~\ref{subsec:hc_curricula} which are summarised in Table~\ref{tab:hc_curriculum_mrpc}.
Ultimately, we disclose the hardware infrastructure that we used to conduct all experiments (Appendix~\ref{app:comp_resources}).

\section{Hyperparameter details}
\label{app:hyperparams}
Throughout our experiments, we employed different sets of hyperparameters. 
In the following tables, we summarise the hyperparameter settings for every experiment, separated by hyperparameters for training and fine-tuning, for model architectures \citep[if not given by][]{raghu2020teaching} and for the schedule functions of our hand-crafted curricula:

\subsection{Hyperparameters training}
\label{app:hyperparams_training}
Here, `variable' values are set depending on the specific subset of GLUE we train on. RoBERTa-models that were trained with hand-crafted (HC) curricula were trained using suboptimal (LOW) and optimal (OPT) learning rates.
\begin{table}[h!]
\caption{Hyperparameters training.}
\label{tab:hyperparameters_training}
\begin{center}
\begin{small}
\begin{sc}
\resizebox{\textwidth}{!}{
\begin{tabular}{lllccccccr}
\toprule
Experiment & & $\gamma$ (lr) & lr-decay & batch size & warm-up & epochs & $I_{practice}$ & $I_{teacher}$   \\
\midrule
 \S~\ref{subsec:comm_exp_setup}: Commentaries 
 & CIFAR ($T$)  & inner: $10^{-4}$ ; outer: $10^{-3}$  &  None  & 8 & - & - & 1500 & 100 \\
 & GLUE ($T$)   & inner: $10^{-4}$; outer: $10^{-3}$ & None & 8 & - & - & variable & 100  \\
 & CIFAR ($S$)  & 2l-CNN: $10^{-3}$; ResNet: $10^{-5}$  & None & 64 & - & 25 & - & -   \\
 & GLUE ($S$)   & RoBERTa: $4 \times 10^{-6}$  & square-root & 8 & 100 & variable & - & - \\

\midrule

\S~\ref{para:experiments_comms}: HC curricula & all   & low: $4 \times 10^{-6}$; opt.: $2 \times 10^{-5}$  &  square-root & 8 & 100 & variable & - & - \\
\bottomrule
\end{tabular}
}
\end{sc}
\end{small}
\end{center}

\end{table}

\subsection{Hyperparameters model architecture}
\label{app:hyperparams_models}

 For all replications of \citet{raghu2020teaching}'s experiments, we used their exact same model architectures. To transfer commentaries to NLP, we conducted a small hyperparameter search to find the smallest possible model architecture for the practice student $S$\textsubscript{p} and teacher ($T$) model that maintains the capacity to substantially reduce the empirical error on all GLUE-benchmark-tasks. The best model follows the transformer-encoder architecture and is implemented using the fairseq library \citep{vaswani2017attention, ott2019fairseq}. $S$\textsubscript{p} and $T$ are using the same base architecture.

\begin{table}[h!]
\caption{Hyperparameters models.}
\label{tab:hyperparameters_models}
\begin{center}
\begin{small}
\begin{sc}
\resizebox{\textwidth}{!}{
\begin{tabular}{llccccr}
\toprule
Experiment & & n-layers & emb-dims & ffn-emb & attention-heads   \\
\midrule
\S~\ref{subsec:comm_exp_setup}: Commentaries 
 & GLUE ($T$ and $S$\textsubscript{p})   & 2  & 64 & 64 & 8 \\

\bottomrule
\end{tabular}
}
\end{sc}
\end{small}
\end{center}
\end{table}

\subsection{Hyperparameters schedule functions}
\label{app:hyperparams_schedule_functions}
We obtain the exact shape of the manual schedule functions from Section~\ref{subsec:hc_curricula} through a hyperparameter grid search and selected the triples in Table~\ref{tab:hyperparameters_schedule_functions} as best performing schedule functions for our two hand-crafted curricula. `Start portion' describes percentage of initially used data, `step size' how much data is added to the portion of used data at every increment and `increment' is the number of updates after which additional data is added to the pool of used data.
\begin{table}[h!]
\caption{Hyperparameters schedule functions.}
\label{tab:hyperparameters_schedule_functions}
\begin{center}
\begin{small}
\begin{sc}
\begin{tabular}{llcccr}
\toprule
Experiment & & start portion & step size & increment    \\
\midrule

\S~\ref{subsec:hc_curricula}: Hand-crafted curricula & Sequence length curriculum   & 30\% &  10\% & 300  \\
 & Loss curriculum  & 30\% &  10\% & 50  \\
  
\bottomrule
\end{tabular}
\end{sc}
\end{small}
\end{center}
\end{table}

\section{Replication commentaries curriculum CIFAR10/100}
\label{app:replication}
\begin{figure*}[!h]
\centering
\begin{subfigure}{\dimexpr0.30\textwidth+20pt\relax}
    \makebox[20pt]{\raisebox{40pt}{\rotatebox[origin=c]{90}{CIFAR10}}}%
    \includegraphics[width=\dimexpr\linewidth-20pt\relax]
    {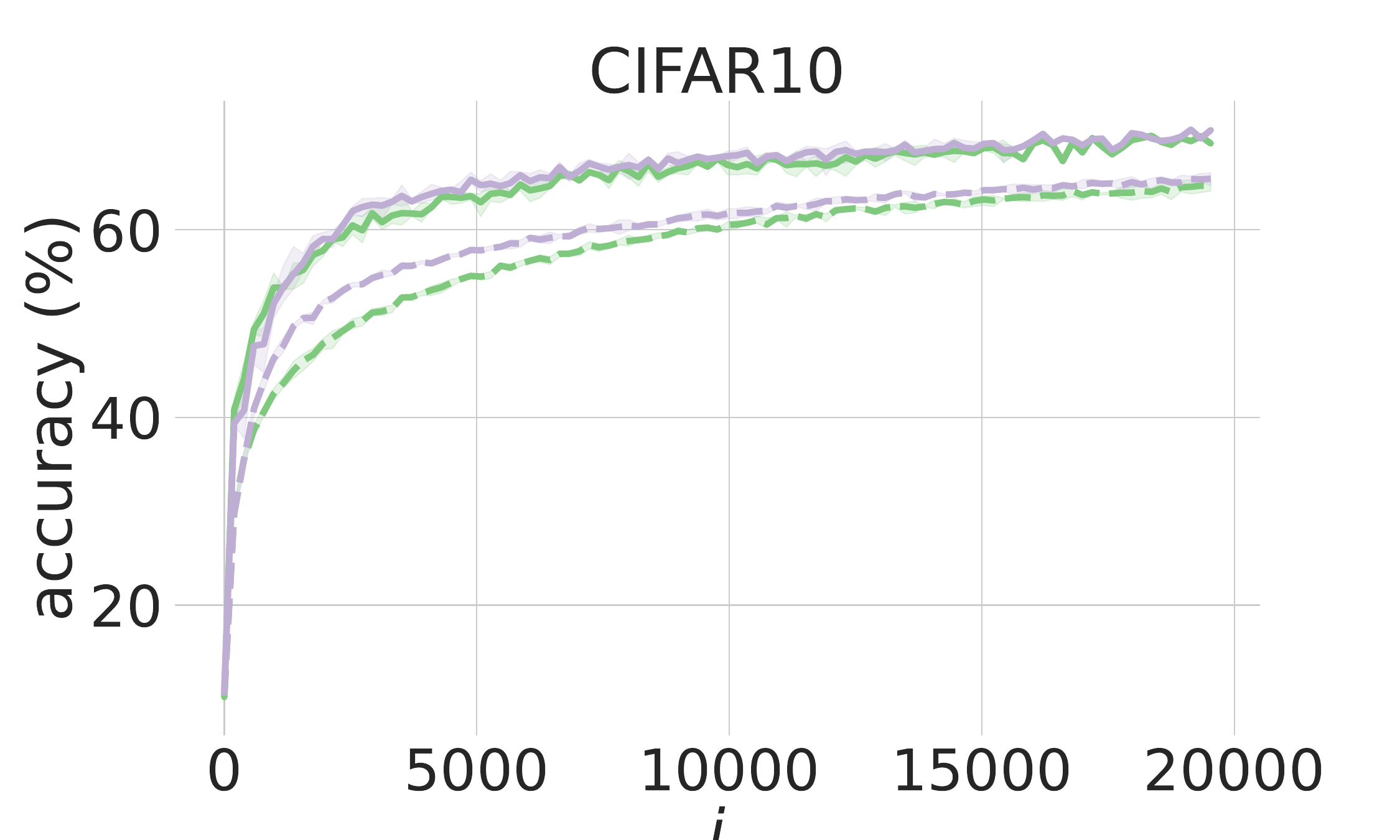}
    \makebox[20pt]{\raisebox{40pt}{\rotatebox[origin=c]{90}{CIFAR100}}}%
    \includegraphics[width=\dimexpr\linewidth-20pt\relax]{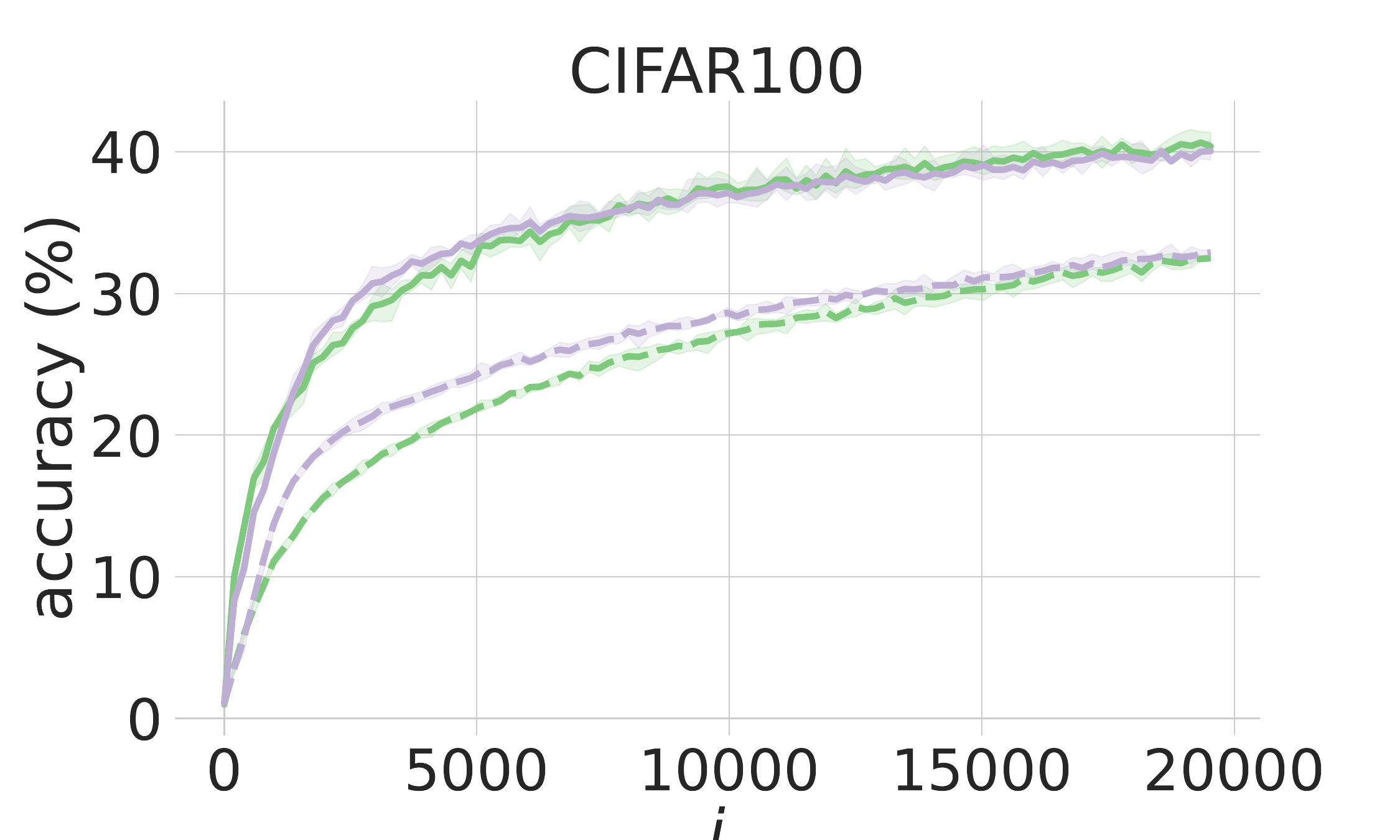}
    \caption{2-layer CNN}
\end{subfigure}\hfill
\begin{subfigure}{0.30\textwidth}
    \includegraphics[width=\textwidth]  
    {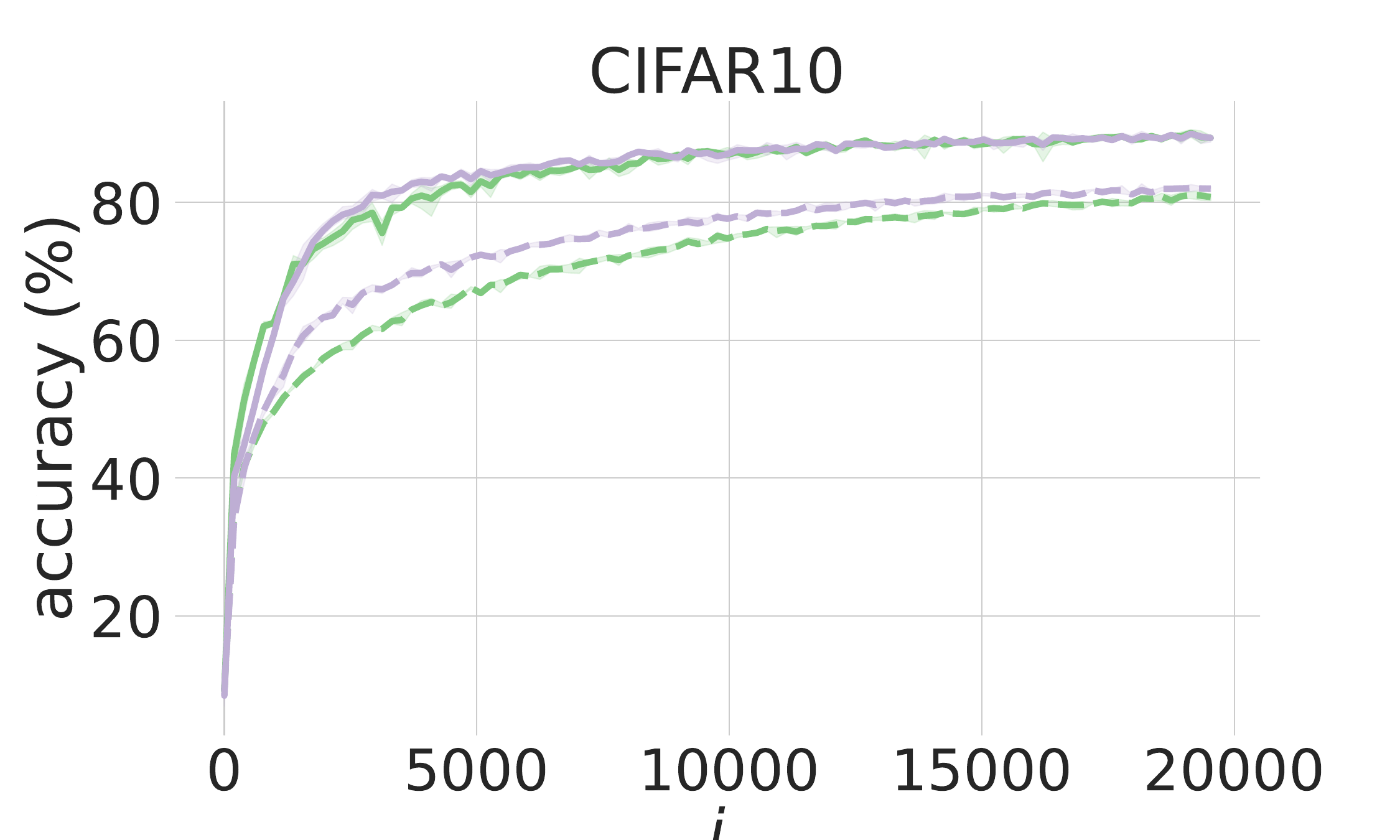}
    \includegraphics[width=\textwidth]{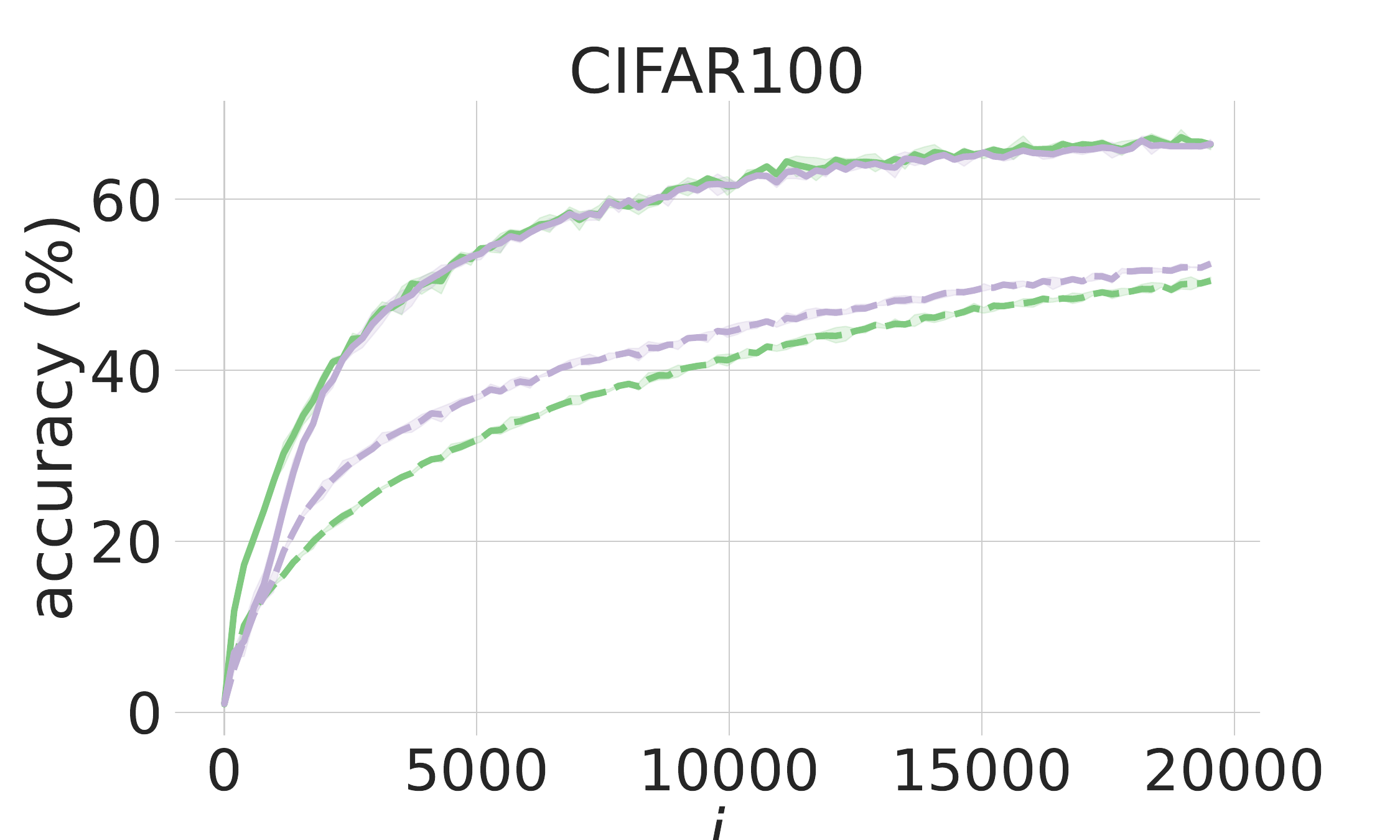}
    \caption{resnet18}
\end{subfigure}\hfill
\begin{subfigure}{0.30\textwidth}
    \includegraphics[width=\textwidth]{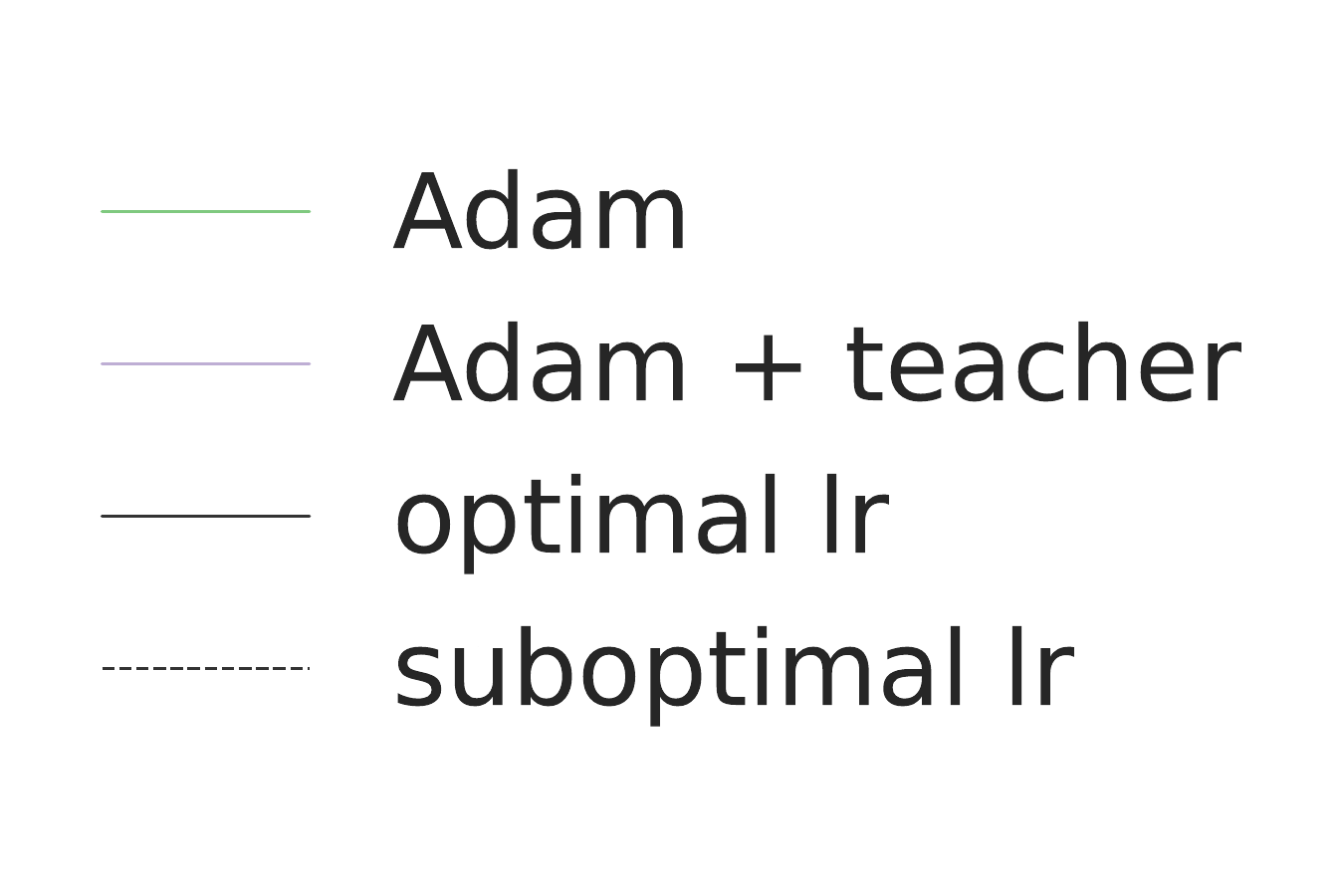}
    \includegraphics[width=\textwidth]{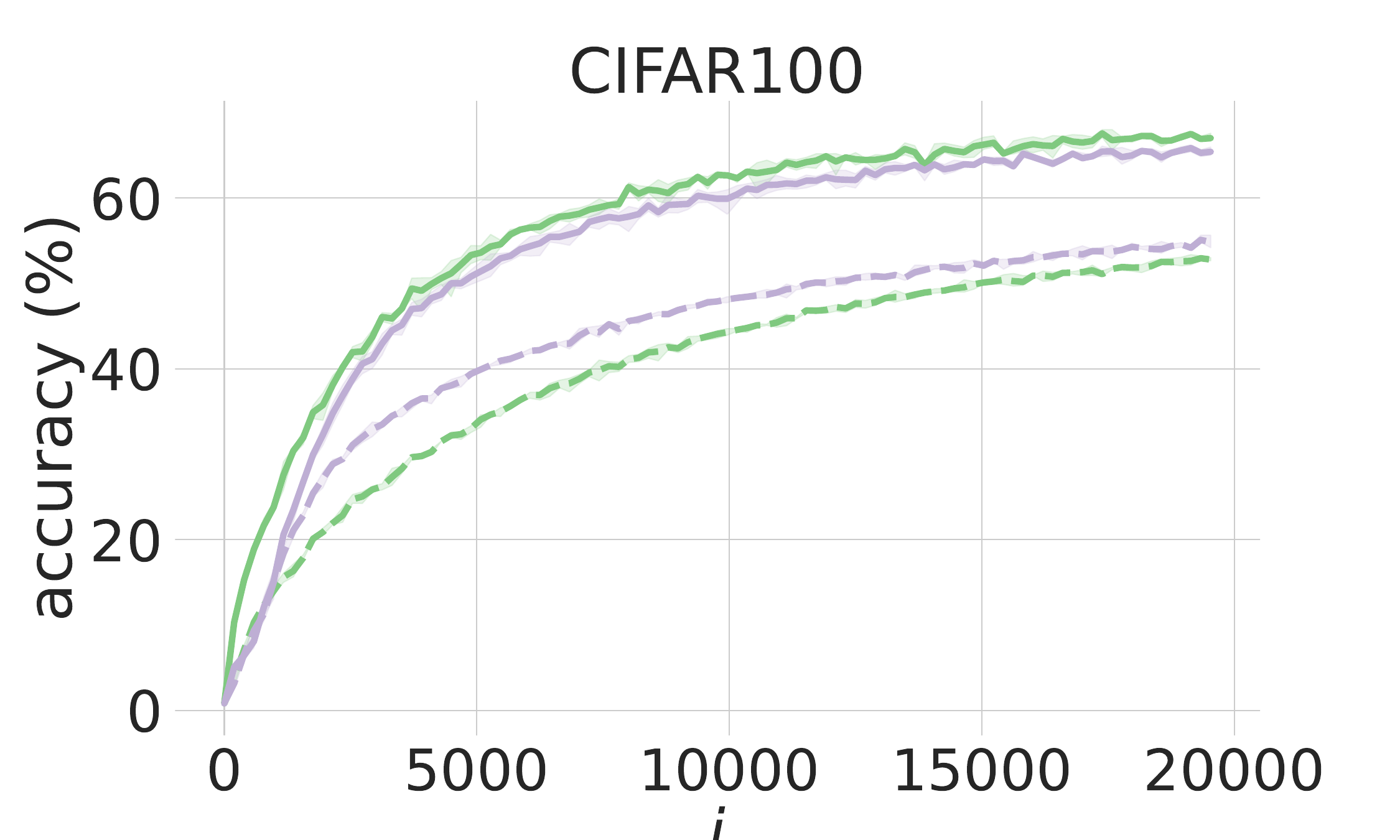}
    \caption{resnet34}
\end{subfigure}
\caption{All replication results from the original paper, with suboptimal hyperparameters that show the effect from the original paper and optimised hyperparamters.}
\label{app_fig:comm_replication}
\end{figure*}

The following (Figure~\ref{app_fig:comm_replication}) shows the performance of different models on CIFAR10 and CIFAR100 when trained with and without the commentaries curriculum. 
We replicate \citet{raghu2020teaching}'s results. 
However, we also find that their hyperparameter setting is suboptimal and that with properly tuned hyperparameters, vanilla Adam outperforms the curriculum.


\begin{figure}[h!]
\centering
\begin{subfigure}{0.48\linewidth}
    \includegraphics[width=\linewidth]{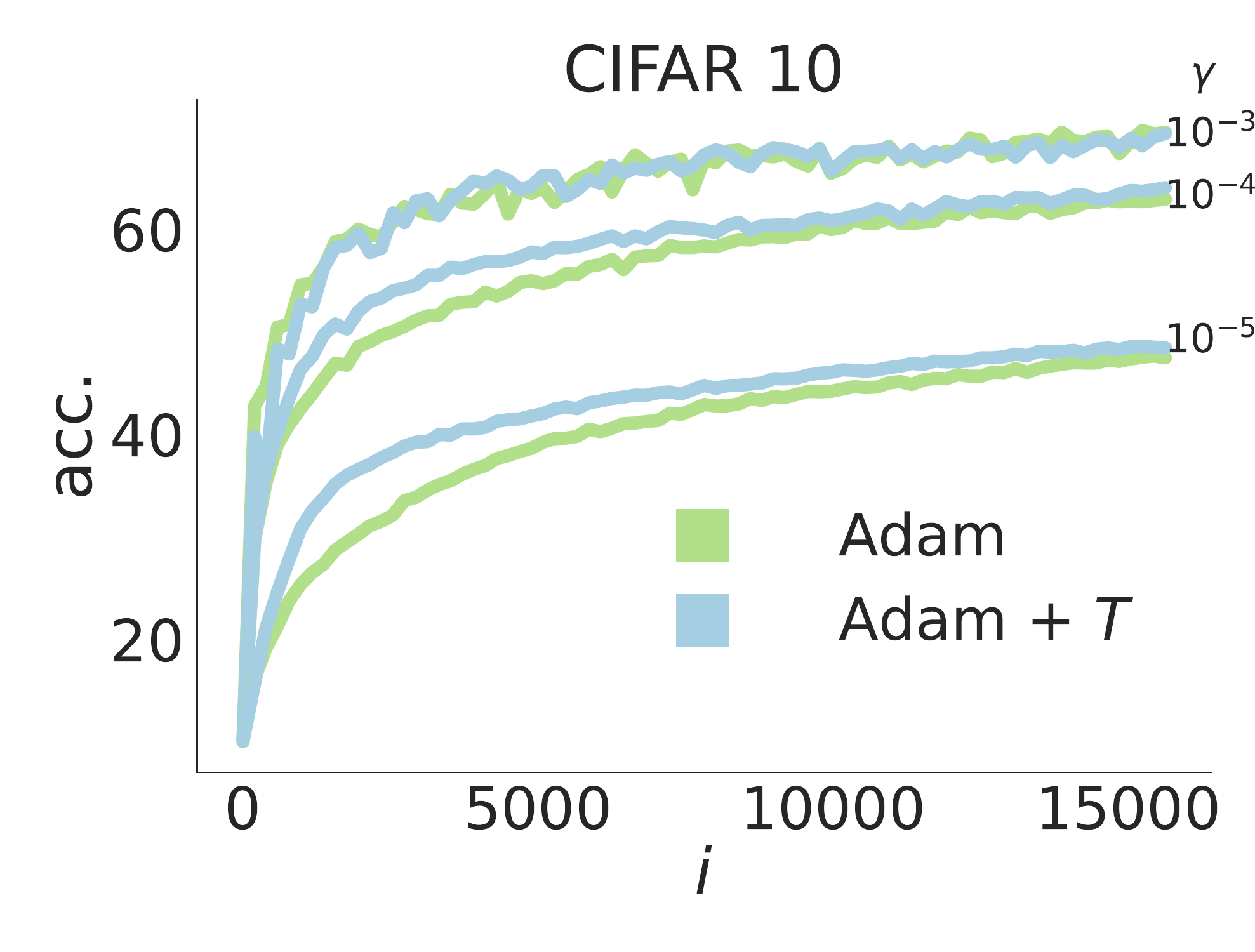}
    \label{subfig:CIFAR_higher_lr}
  \end{subfigure}
  \caption{Learning curves for the 2-layer CNN trained on CIFAR10 with and without teacher at different learning rates $\gamma$. We see how lowering $\gamma$ helps commentaries improve over the vanilla Adam. At the overall best $\gamma$, however, vanilla Adam performs on par.
 }
  \label{fig:CIFAR_higher_lr}
\end{figure}

\FloatBarrier

\section{Toy curricula CIFAR10}
\label{app:toy_teachers}
In this section, we exemplify the simple loss-weighting policies that we described in Section~\ref{subsec:comm_exp_results}.
When applied to the 2-layer CNN model while training on the CIFAR10 dataset, the toy-teacher show how a simple shift of loss-reweighting from low- to high weight values can improve learning speed above no weighting (baseline with $w_i = 1$). We can also see, how decreasing weights have the opposite effect (see $T_\downarrow\textsubscript{linear}$) and that the absolute value of the weight has no influence (compare $T_\textsubscript{constant}$ and baseline).

\begin{figure*}[h!]
  \centering
  \begin{subfigure}{0.40\linewidth}
    \includegraphics[width=\linewidth]{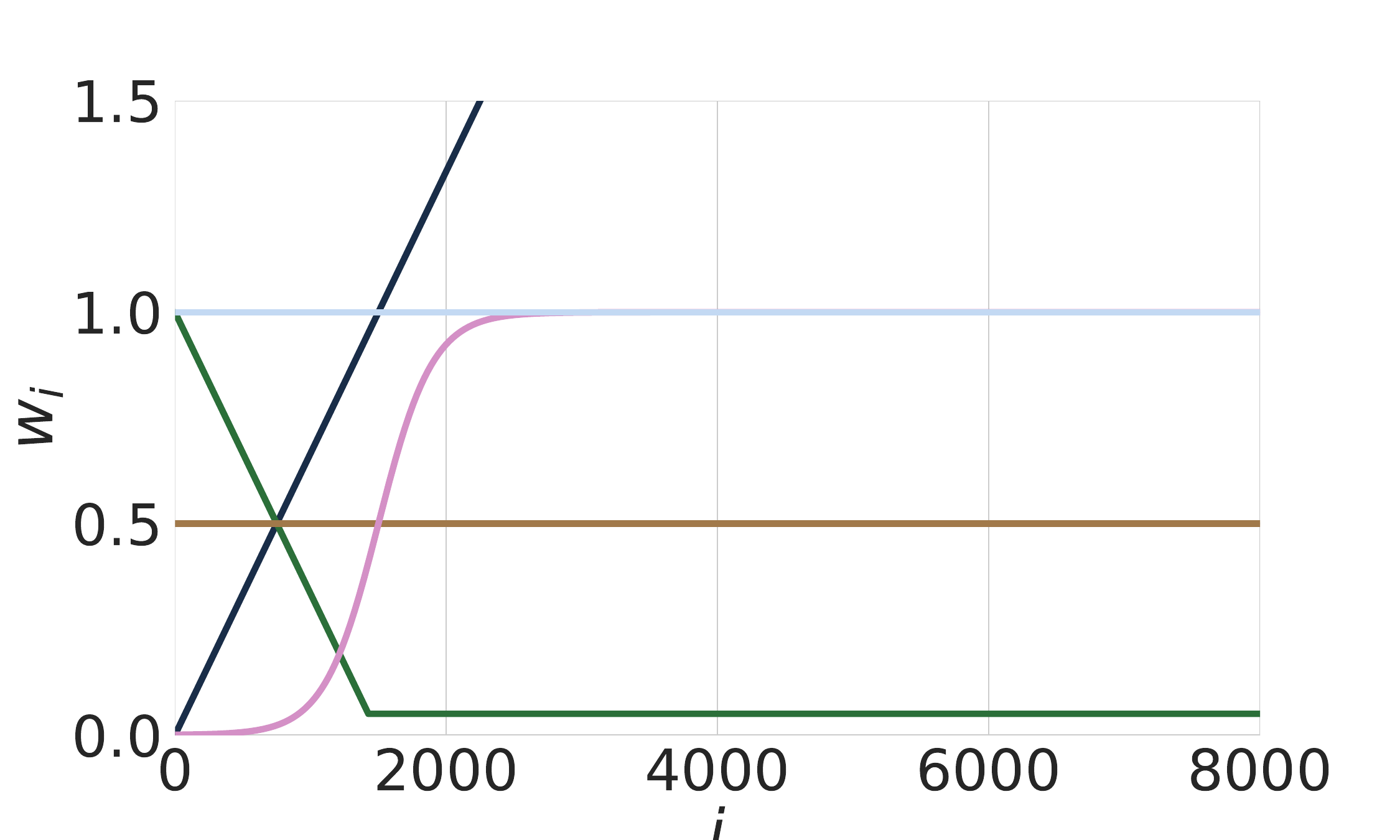}
    \caption{}
  \end{subfigure}
  \begin{subfigure}{0.40\linewidth}
    \includegraphics[width=\linewidth]{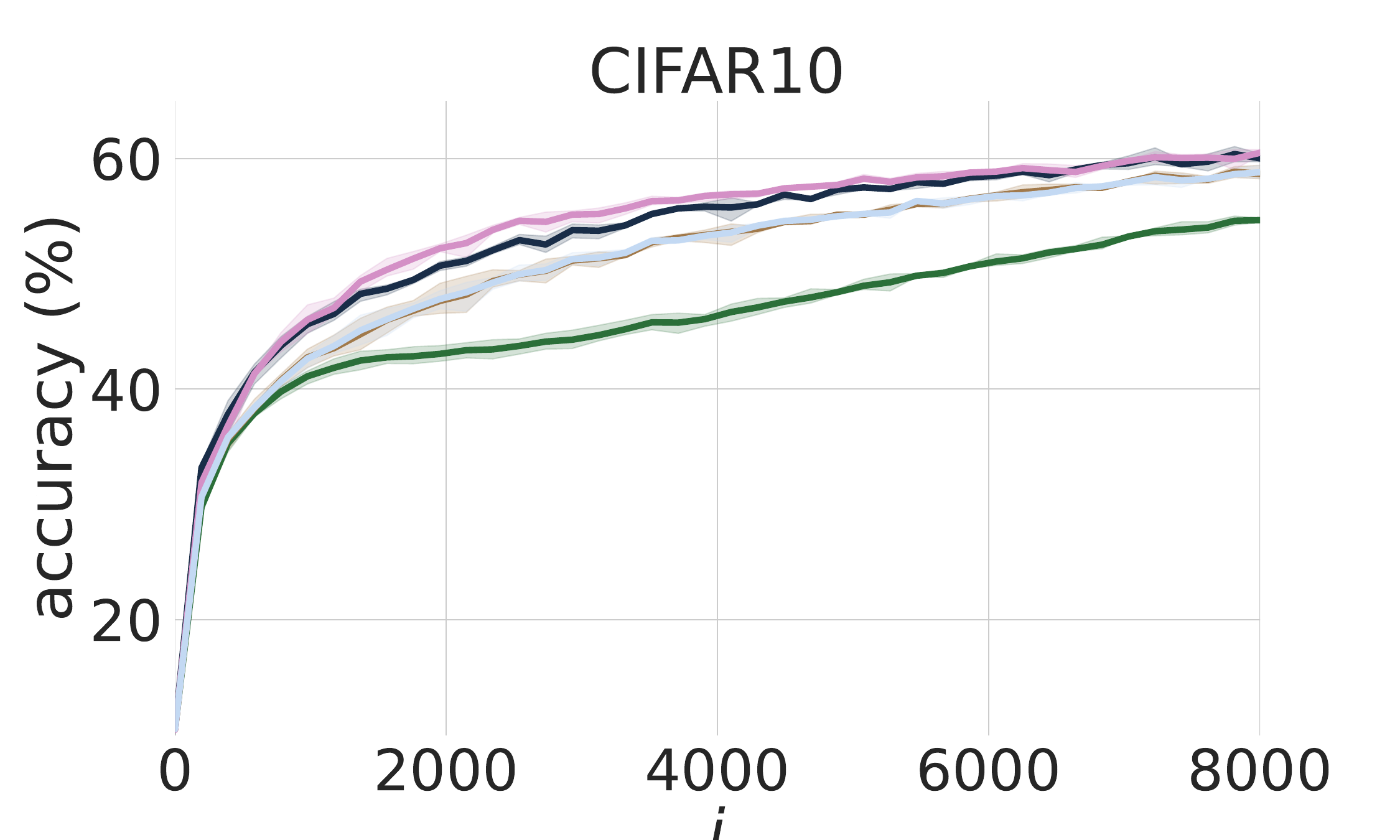}
    \caption{}
  \end{subfigure}
  \begin{subfigure}{0.19\linewidth}
    \includegraphics[width=\linewidth]{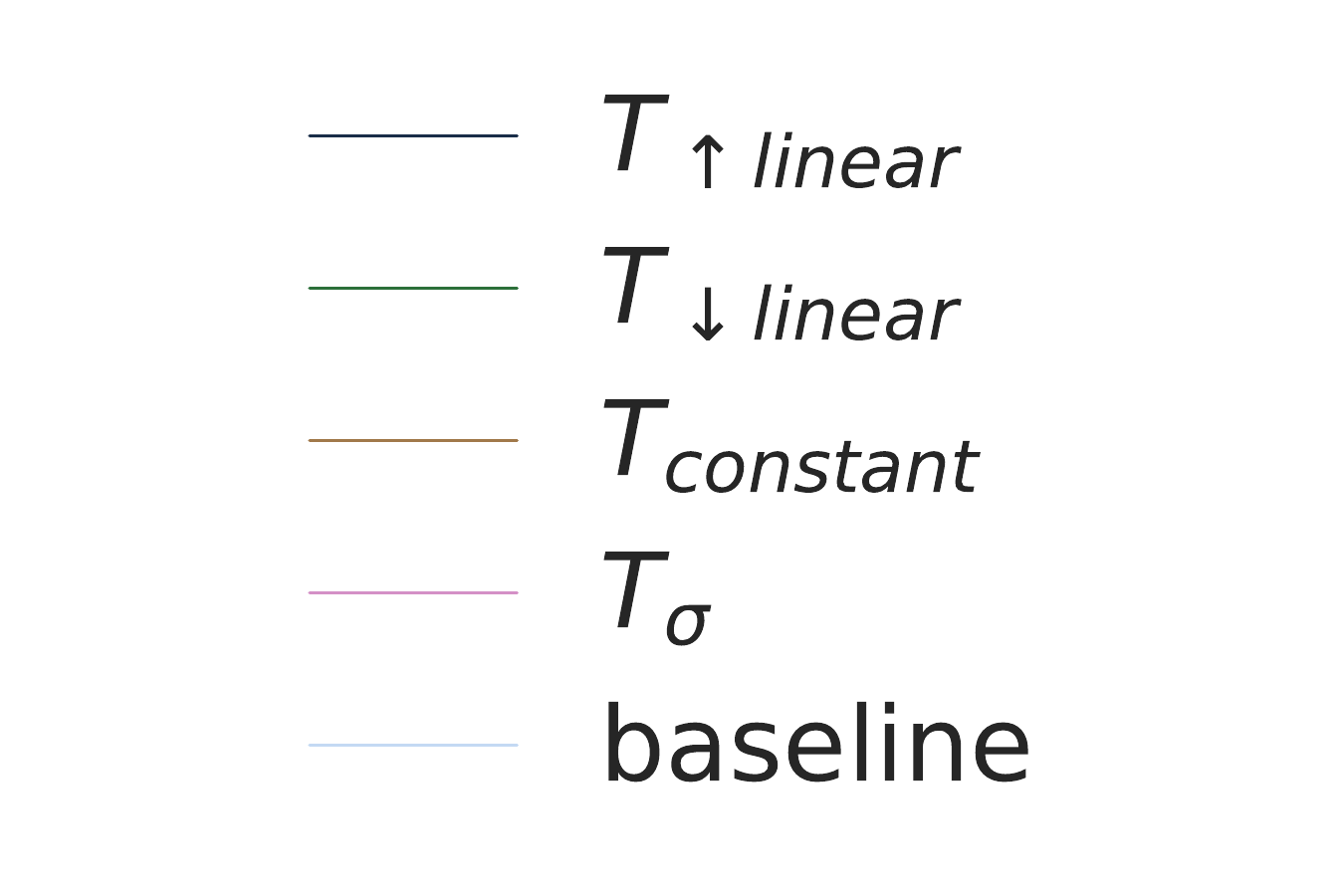}
    \caption{}
  \end{subfigure}
  \caption{The left side (a) shows the weights applied to the loss by the different toy curricula. The right side (b) shows the performance of a 2-layer CNN trained on CIFAR10 with the different toy curricula.}
  \label{app_fig:}
\end{figure*}

\FloatBarrier

\section{GLUE with Commentaries}
\label{app:extension_NLP}
In this section, we document the learning speed improvements that we observe with commentaries when we finetune RoBERTa on different GLUE-tasks. Either axis shows the steps that the models requires to converge to 98\% of its final performance when it is trained with and without a commentaries teacher. We can see how with a suboptimal learning-rate (lr), RoBERTa generally converges faster when it is trained with commentaries (dots land above the diagonal). As soon as we use the optimal learning rate, Adam without a teacher converges faster or just as fast as with teacher (crosses land below the diagonal or on it). 

\begin{figure*}[h!]
  \centering
  \begin{subfigure}{0.60\linewidth}
    \includegraphics[width=\linewidth]{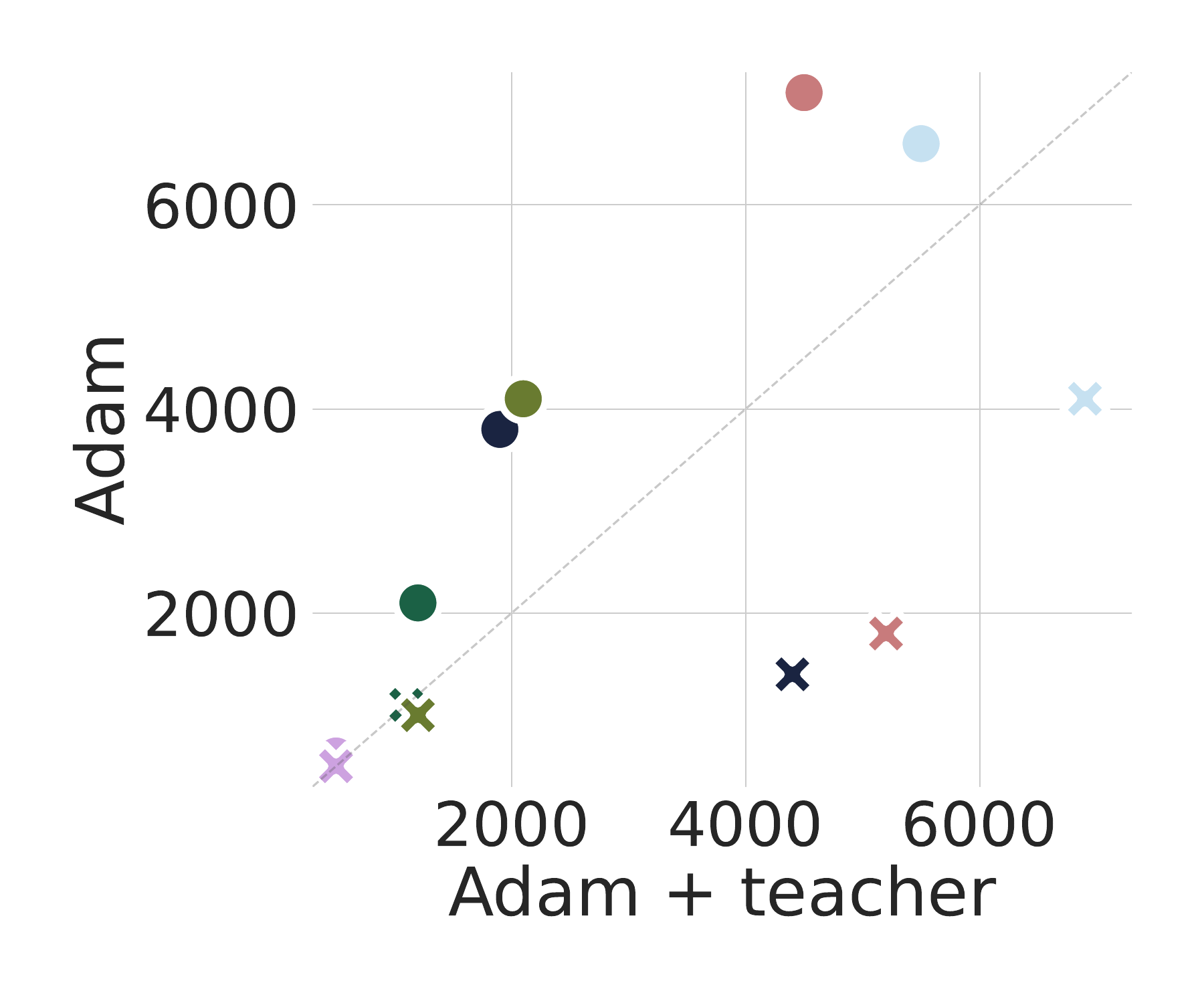}
    \label{app_fig:commentaries_glue_low_LR}
  \end{subfigure}
\begin{subfigure}{0.19\linewidth}
    \includegraphics[width=\linewidth]{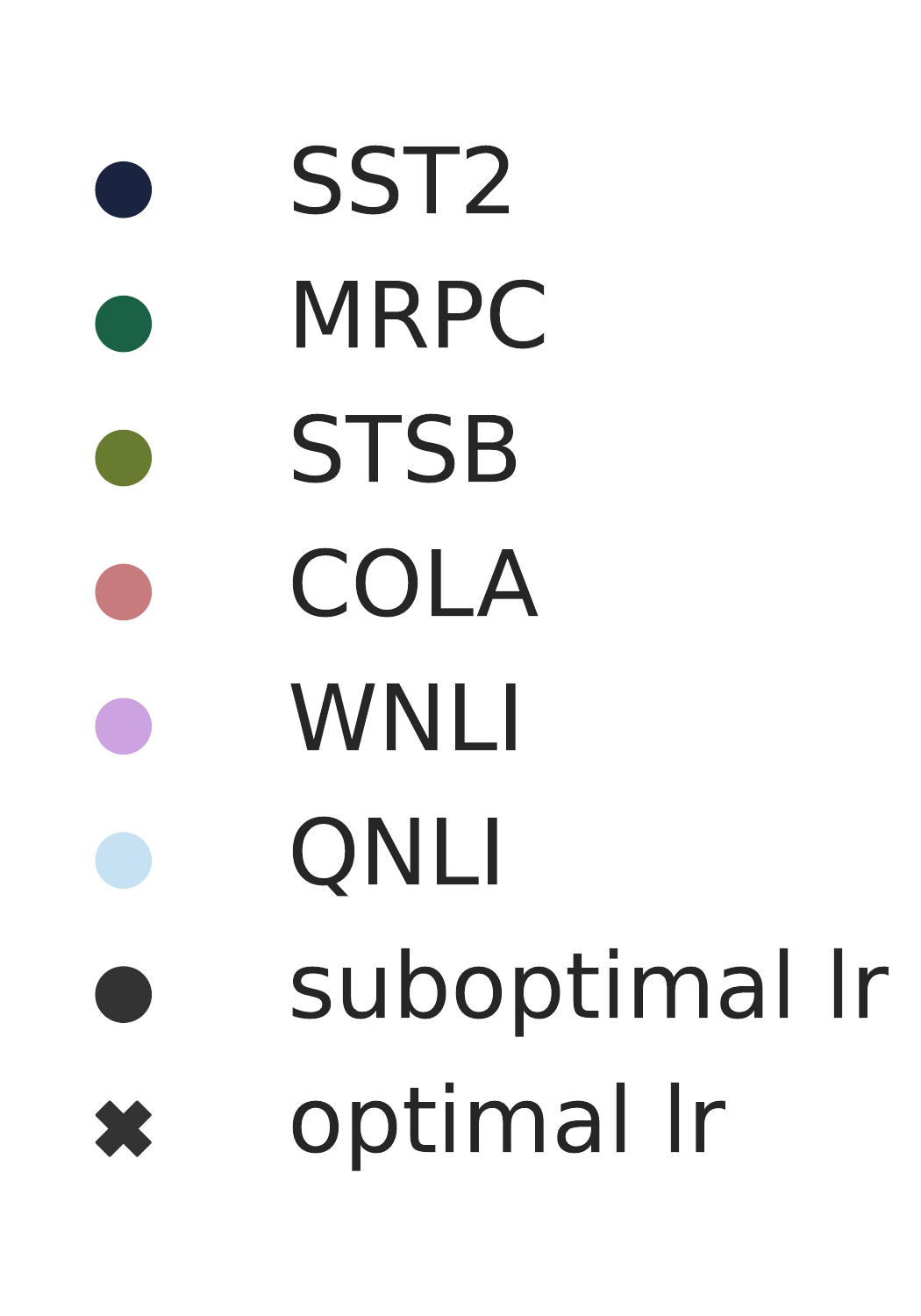}
    \label{app_fig:commentaries_glue_high_LR}
  \end{subfigure}
  \caption{Updates RoBERTa\textsubscript{BASE} needs to converge when finetuned on different GLUE tasks, with and without teacher. Dots above the line mean that the model with teacher learns faster; dots below the line mean the model without teacher is faster. We see how an optimal learning rate eliminates the effects of the teacher. Convergence is defined as 98\% of final validation performance.}
  \label{app_fig:commentaries_glue}
\end{figure*}

\FloatBarrier

\section{Correlations of difficulty measures with $|g|$}
\label{app:correlations_diffm_and_gradients}
We stated in Section~\ref{subsec:hc_curricula} that difficulty measures are correlated with the size of the gradient that they evoke in a model. 
We here show empirically that this is the case for the two difficulty measures that we are considering in our experiments (sequence length and loss).

\begin{figure}[h!]
  \centering
  \begin{subfigure}{0.49\linewidth}
    \includegraphics[width=\linewidth]{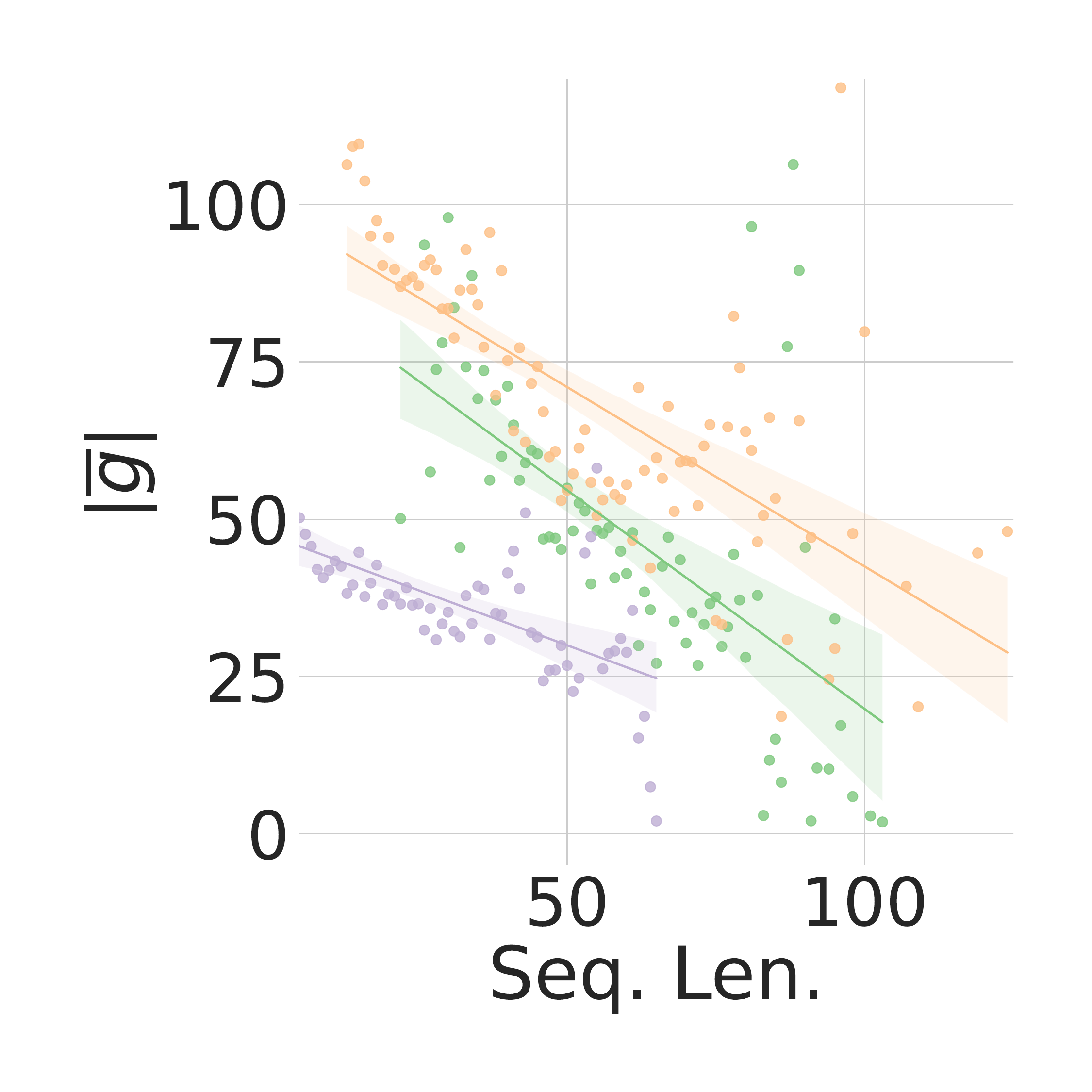}
    \caption{}
    \label{subfig:scatter_grad_seq_len}
  \end{subfigure}
  \begin{subfigure}{0.49\linewidth}
    \includegraphics[width=\linewidth]{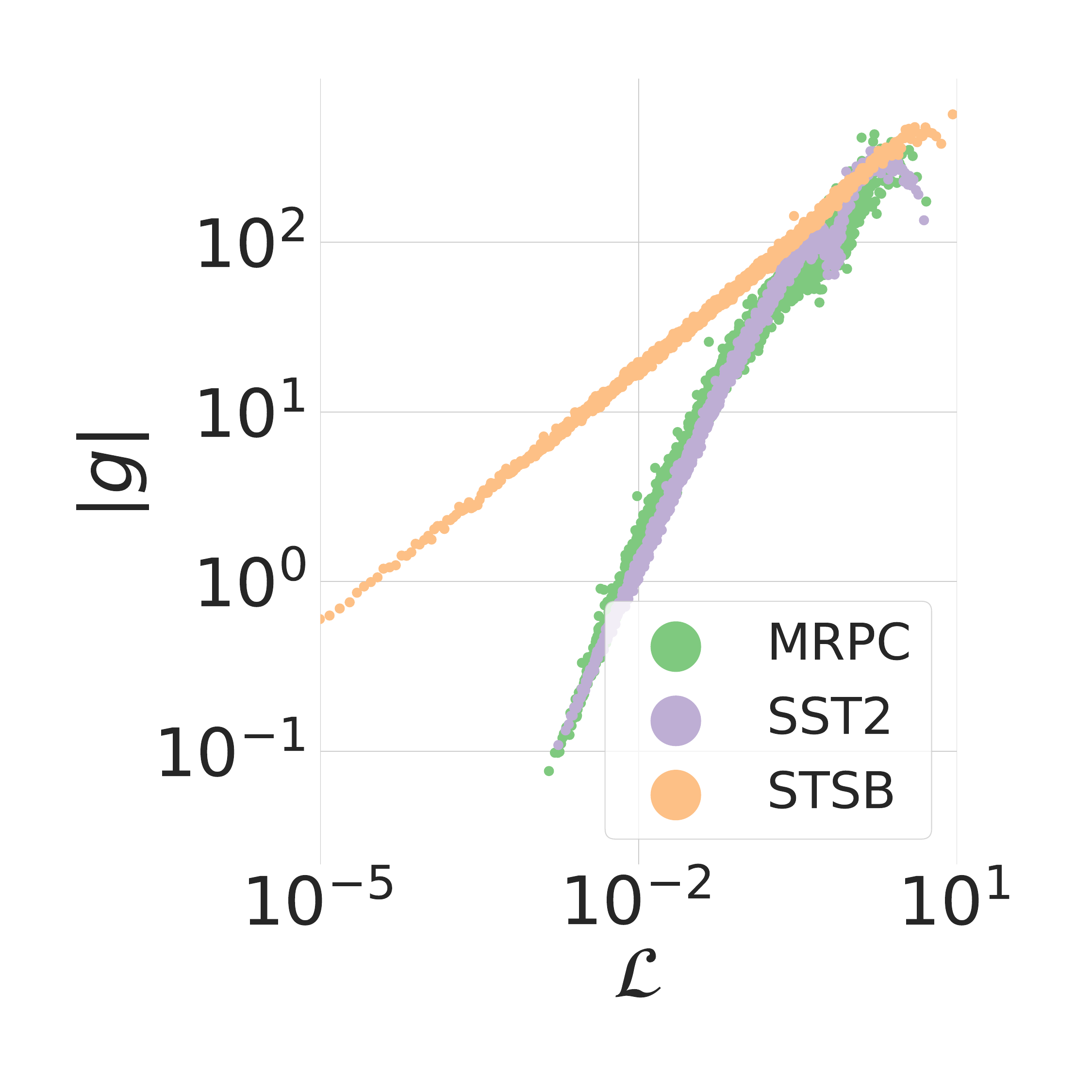}
    \caption{}
    \label{subfig:scatter_grad_loss}
  \end{subfigure}
  \caption{Covariance of common difficulty measures (Sequence length and Loss) with the size of gradients that they produce when fine-tuning RoBERTa\textsubscript{BASE} for a selection of GLUE-tasks. 
  Both, sequence lengths (a) and by cross-entropy-loss (b) are highly correlated with the average gradient norms. We chose a representative subset of GLUE and binned data points to improve the presentability of the results.}
  \label{fig:scatter}
\end{figure}

\FloatBarrier

\section{Learning curves hand-crafted curricula}
\label{app:learning_curves_hc}
In this section, we present the learning-curves that correspond to the training-runs summarised in Table~\ref{tab:hc_curriculum_mrpc}. We can see that our hand-crafted curricula only provide an advantage when $\gamma$ is set low. As soon as we use an optimal learning rate, plain Adam outperforms the curricula. Moreover, learning with the curricula becomes highly unstable (see by variance across runs), something that is generally known to happen when parameter updates are too large. Ultimately, we can also see how the benefit in hand-crafted curricula can also be eliminated by setting beta-values to equal values, just like we previously observed it for commentaries before.
\begin{figure*}[h!]
  \centering
  \begin{subfigure}{0.32\linewidth}
    \includegraphics[width=\linewidth]{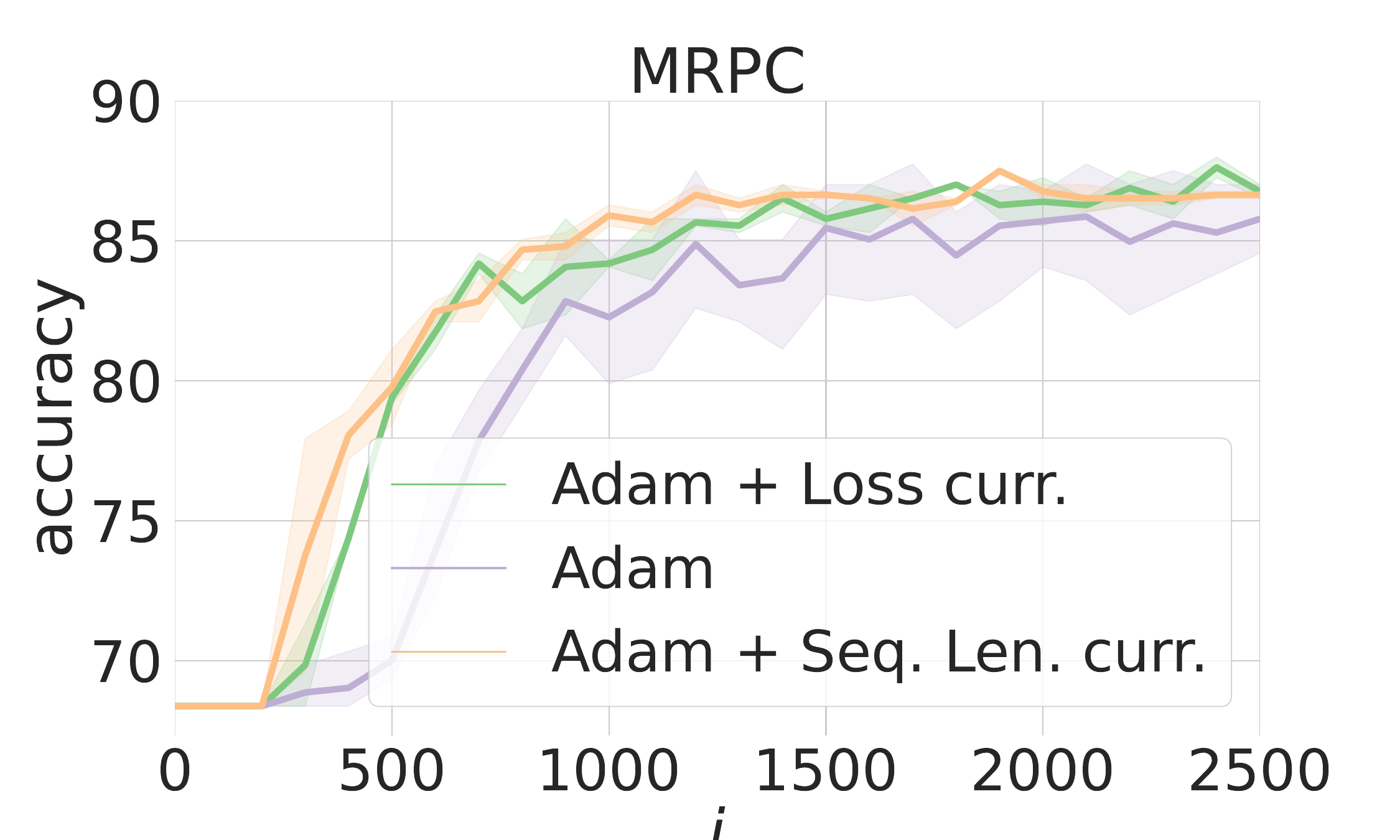}
    \caption{default $\beta$-hyperparameters + low $\gamma$}
    \label{subfig:predefined_curricula_perf_low_lr}
  \end{subfigure}
    \begin{subfigure}{0.32\linewidth}
    \includegraphics[width=\linewidth]{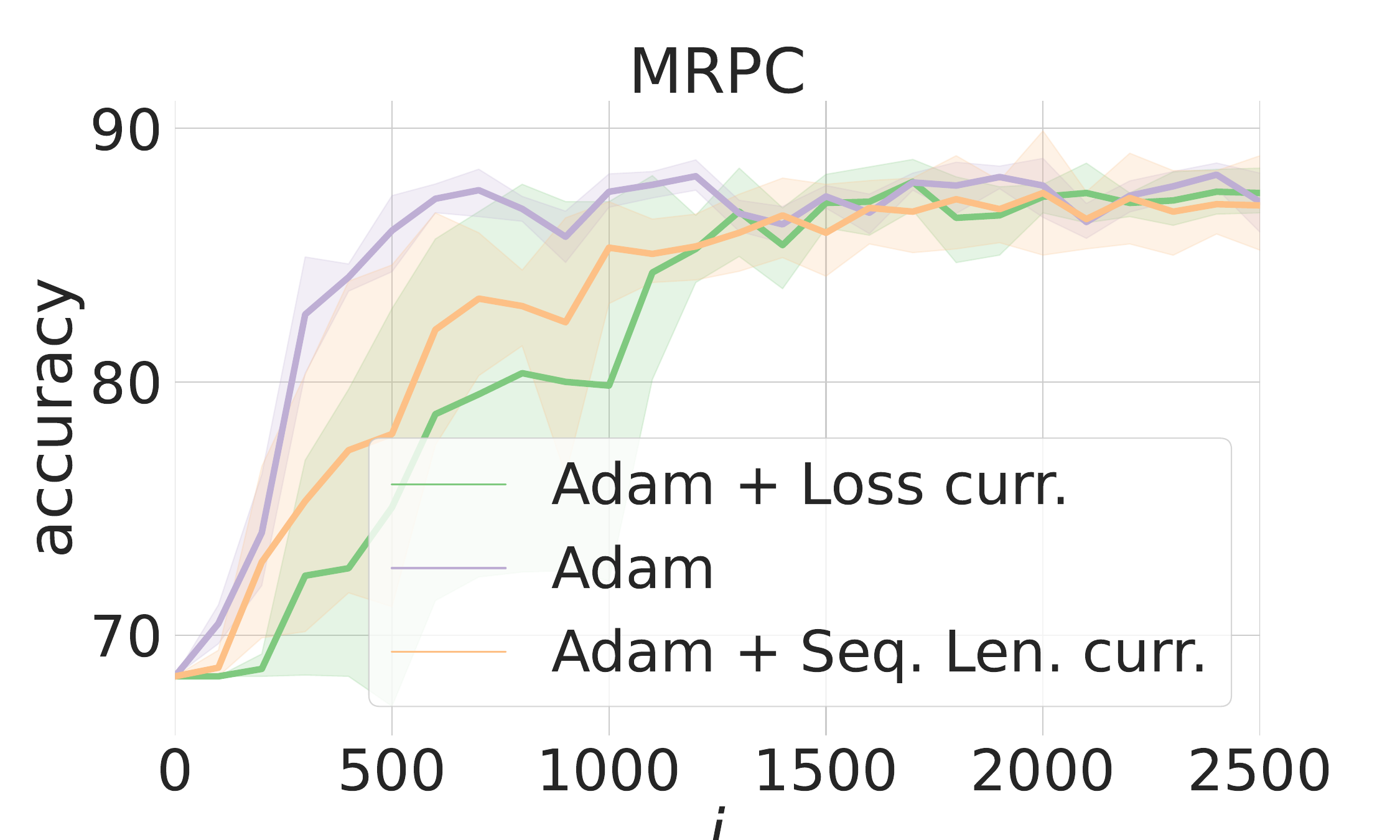}
    \caption{default $\beta$-hyperparameters + high $\gamma$}
    \label{subfig:predefined_curricula_perf_high_lr}
  \end{subfigure}
  \begin{subfigure}{0.32\linewidth}
    \includegraphics[width=\linewidth]{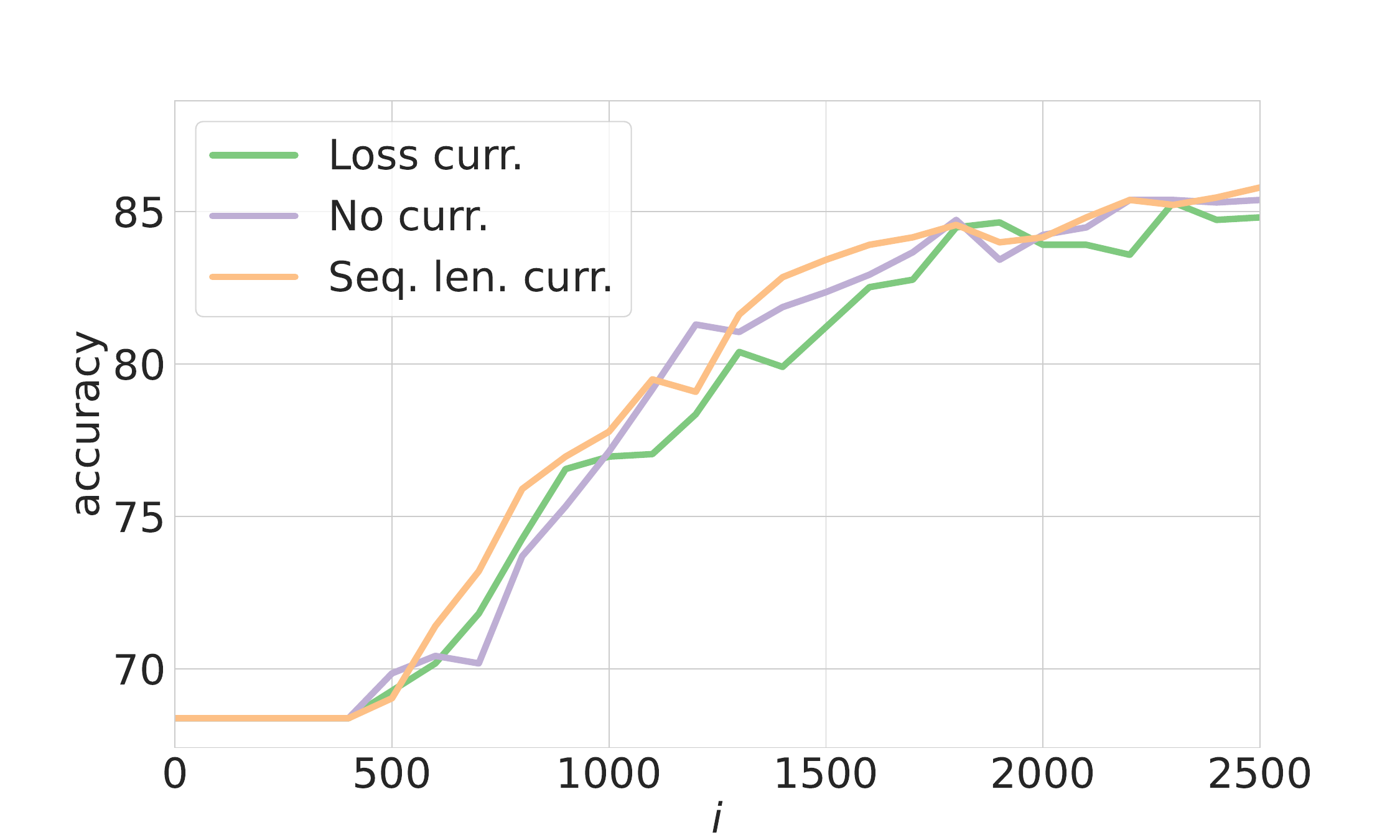}
    \caption{$\beta_1$ = $\beta_2$ = 0.99}
    \label{subfig:predefined_curricula_perf_beta_equal}
  \end{subfigure}
  \caption{Learning curves of RoBERTa\textsubscript{BASE} when finetuned on MRPC trained with the hand-crafted curricula. (a) shows the performance when Adam's $\beta$-parameters allow for interaction. The learning rate $\gamma$ = 4e-6 lets our hand-crafted curricula outperform the baseline using vanilla Adam. (b) Shows what happens with optimal $\gamma$ = 2e-5: vanilla Adam outperforms any curriculum condition. Additionally, the curricula conditions become unreliable across runs, visible in the shaded area of the confidence interval. (c) shows the performance when interactions are prevented. Here, the curricula do not yield any learning advantage.
  }
  \label{fig:predefined_curricula_perf}
\end{figure*}

\FloatBarrier

\section{Computational resources}
\label{app:comp_resources}
In this very last section, we disclose the computational infrastructure that was necessary to conduct our experiments. As commentaries require to save the whole computational graph of the practice student's training to be saved, GPUs with larger vRAM are desirable. 

\begin{table}[h!]
\caption{Computational resources used for conducting our experiments.}
\label{sample-table}
\begin{center}
\begin{small}
\begin{sc}
\begin{tabular}{lcccr}
\toprule
Resources & Type & Quantity & Capacity  \\
\midrule
GPUs   & NVIDIA A30  & 5 &  24GB HBM2    \\
CPUs   & Intel Xeon Silver &  25 &  2.4GHz x 10 \\
RAM   & -- & 1 & 256GB \\
\bottomrule
\end{tabular}
\end{sc}
\end{small}
\end{center}
\end{table}


\end{document}